\documentclass[11pt]{article}

\usepackage[preprint]{acl}

\usepackage{times}
\usepackage{latexsym}

\usepackage[T1]{fontenc}

\usepackage[utf8]{inputenc}

\usepackage{microtype}

\usepackage{inconsolata}

\usepackage{graphicx}

\usepackage{svg}
\usepackage{multirow}
\usepackage{caption}
\usepackage{amsmath,hyperref}

\usepackage{booktabs}
\usepackage{arydshln}
\usepackage{makecell}
\usepackage{array}
\usepackage{adjustbox}
\usepackage{subcaption}
\usepackage{tikz}
\usepackage[table]{xcolor}
\usepackage{colortbl}
\usepackage{pifont}
\usetikzlibrary{arrows.meta}

\usepackage{mathtools} 

\usepackage{amssymb}

\definecolor{lightgrayrow}{gray}{0.9}

\newcommand\scrolls{\textsc{Scrolls}}

\title{CoMeT: Collaborative Memory Transformer for Efficient Long Context Modeling}

\author{Runsong Zhao\textsuperscript{1,3}\thanks{Equal contribution.} \quad
        Shilei Liu\textsuperscript{3}\footnotemark[1] \quad
        Jiwei Tang\textsuperscript{2,3} \\
        \textbf{Langming Liu}\textsuperscript{3} \quad
        \textbf{Haibin Chen}\textsuperscript{3} \quad
        \textbf{Weidong Zhang\textsuperscript{3}} \quad
        \textbf{Yujin Yuan\textsuperscript{3}} \\
        \textbf{Tong Xiao\textsuperscript{1}\thanks{Corresponding authors.}} \quad
        \textbf{Jingbo Zhu\textsuperscript{1}} \quad
        \textbf{Wenbo Su\textsuperscript{3}} \quad
        \textbf{Bo Zheng\textsuperscript{3}\footnotemark[2]} \\
        \\
        \textsuperscript{1}{\normalsize School of Computer Science and Engineering, Northeastern University, Shenyang 110819, China} \\
        \textsuperscript{2}{\normalsize Tsinghua University} \quad
        \textsuperscript{3}{\normalsize Future Living Lab of Alibaba} \\
        \texttt{zhaors@mails.neu.edu.cn} \quad \texttt{liushilei.lsl@taobao.com}
}

\begin{document}
\maketitle
\begin{abstract}

The quadratic complexity and indefinitely growing key-value (KV) cache of standard Transformers pose a major barrier to long-context processing. To overcome this, we introduce the \textbf{Co}llaborative \textbf{Me}mory \textbf{T}ransformer (CoMeT), a novel architecture that enables LLMs to handle arbitrarily long sequences with constant memory usage and linear time complexity. Designed as an efficient, plug-in module, CoMeT can be integrated into pre-trained models with only minimal fine-tuning. It operates on sequential data chunks, using a dual-memory system to manage context: a temporary memory on a FIFO queue for recent events, and a global memory with a gated update rule for long-range dependencies. These memories then act as a dynamic soft prompt for the next chunk. To enable efficient fine-tuning on extremely long contexts, we introduce a novel layer-level pipeline parallelism strategy. The effectiveness of our approach is remarkable: a model equipped with CoMeT and fine-tuned on 32k contexts can accurately retrieve a passkey from any position within a 1M token sequence. On the \scrolls{} benchmark, CoMeT surpasses other efficient methods and achieves performance comparable to a full-attention baseline on summarization tasks. Its practical effectiveness is further validated on real-world agent and user behavior QA tasks. The code is available at: \url{https://github.com/LivingFutureLab/Comet}

\end{abstract}

\section{Introduction}

\begin{figure*}[t]
\centering
\captionsetup[sub]{font=small, labelfont=small}
 \begin{minipage}{0.5\textwidth}
        \centering
        \subcaptionbox{Passkey Retrieval Accuracy\label{sub1}}{
         \includegraphics[width=\linewidth]{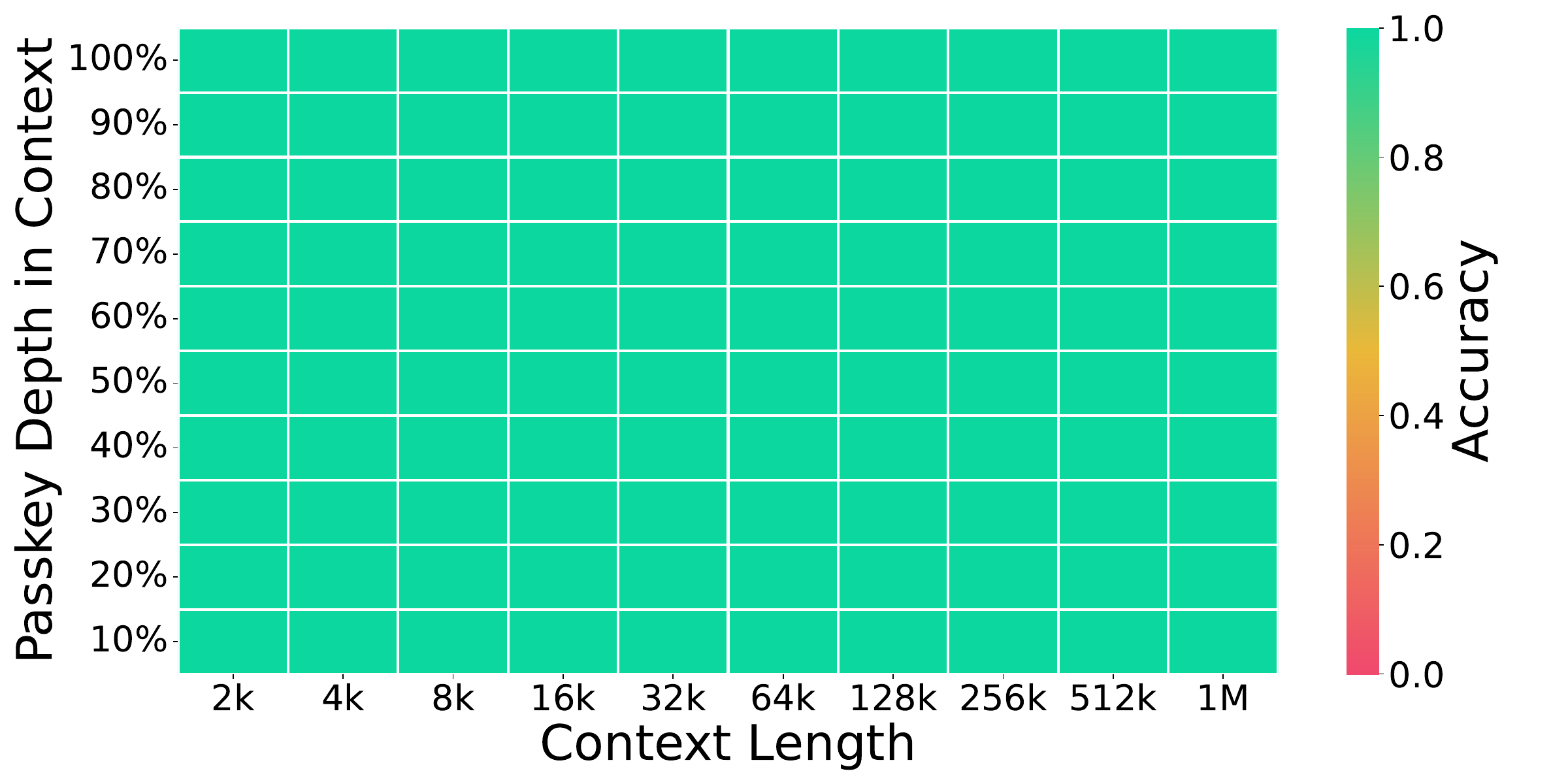} 
        }
 \end{minipage}
 \begin{minipage}{0.25\textwidth}
    \centering
    \subcaptionbox{Inference time\label{sub2}}{ 
    \includegraphics[width=\linewidth]{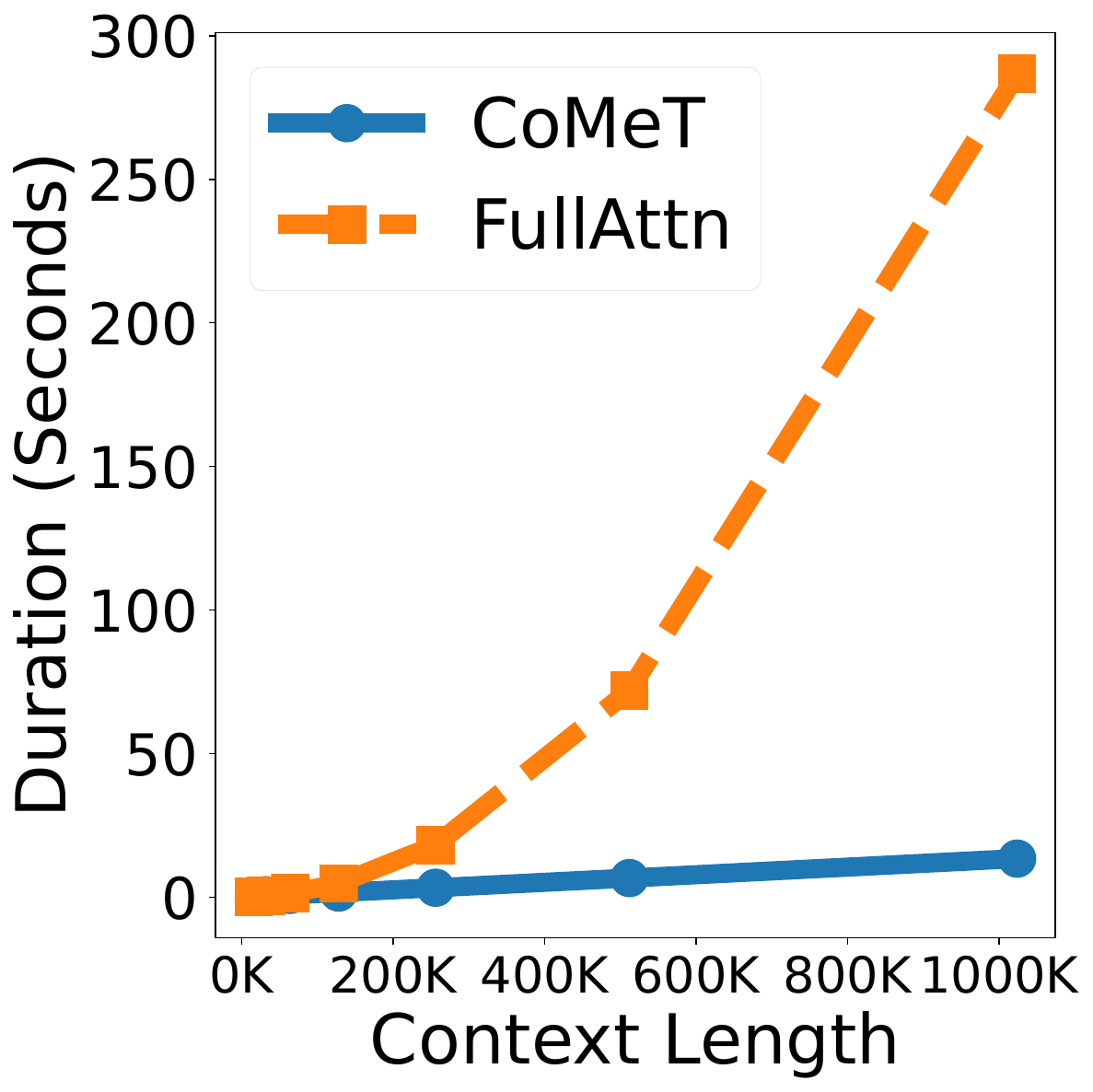}
    }
\end{minipage}
\begin{minipage}{0.25\textwidth}
    \centering
    \subcaptionbox{GPU memory\label{sub3}}{ 
    \includegraphics[width=\linewidth]{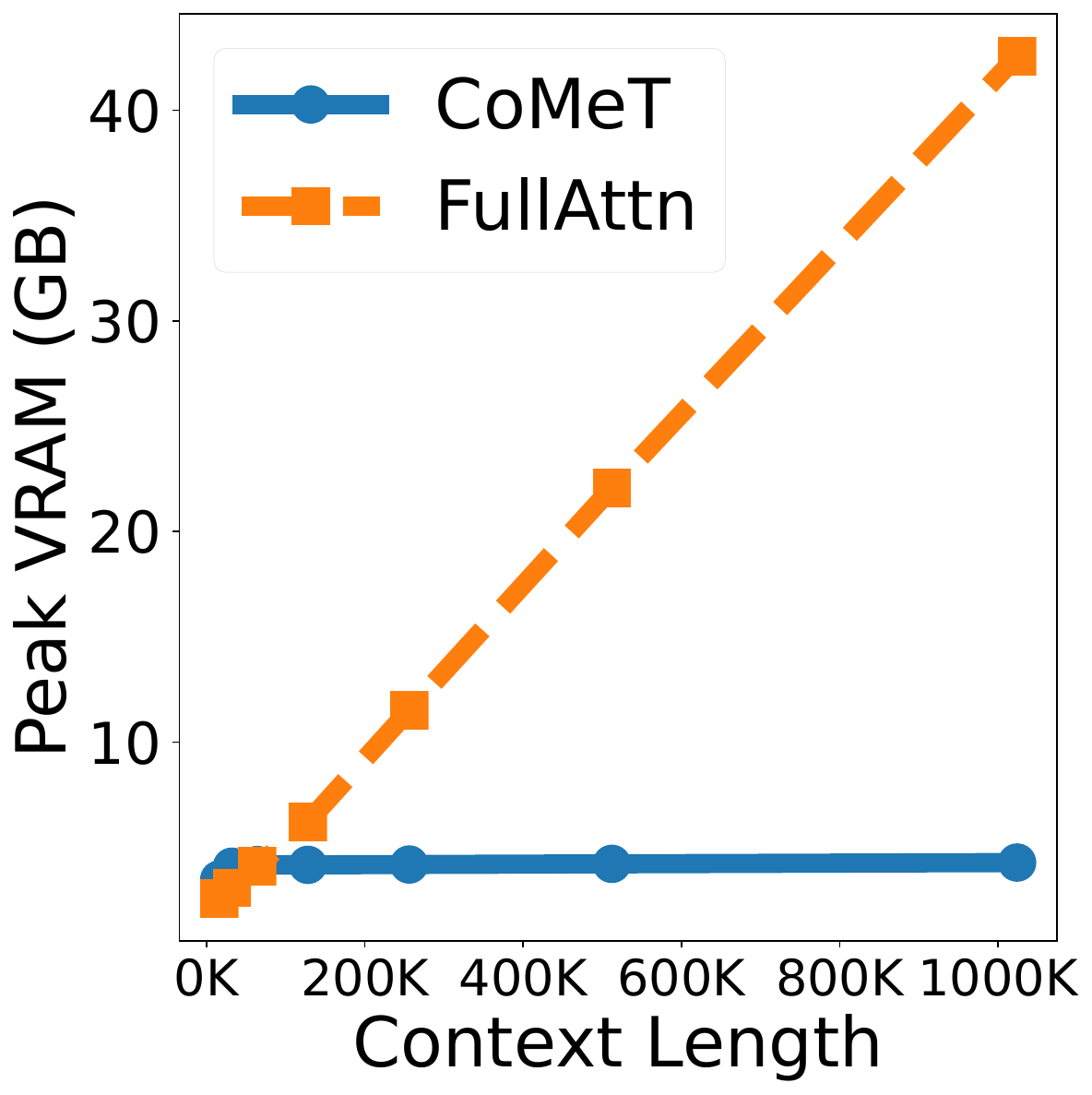}
    }
\end{minipage}
\caption{CoMeT is trained on the passkey task~\cite{munkhdalai2024leave} (i.e., the \emph{Needle-in-a-Haystack} test) with a 32k context, yet it can retrieve a passkey from any position within a 1M-token context. Moreover, its inference time scales linearly with the context length, while GPU memory usage remains constant.}
\label{fig:Main_Fig}
\end{figure*}

The ability to process and reason over vast contexts is a crucial frontier for Large Language Models (LLMs). From processing long documents for summarization~\cite{huang-etal-2021-efficient,pang-etal-2023-long} and question answering~\cite{zhang-etal-2025-longcite,huang-etal-2021-efficient}, to engaging in complex, multi-turn dialogues~\cite{laban2025llmslostmultiturnconversation,yi2024survey} and comprehending large codebases~\cite{yuan2023evaluatinginstructiontunedlargelanguage}, the capacity to capture long-range dependencies is a prerequisite for unlocking the full potential of LLMs in real-world applications. This requires models to not only understand but also persistently retain information across thousands or even millions of tokens, enabling them to grasp intricate narrative structures and make inferences based on a complete history.

However, the architectural foundation of modern LLMs, the Transformer~\cite{vaswani2017attention}, faces a fundamental scaling crisis when confronted with long sequences. Its standard implementation relies on a key-value (KV) cache that grows linearly with the input length, while the attention mechanism incurs quadratic computational complexity (as illustrated in Figures~\ref{sub2} and~\ref{sub3}). This makes processing extremely long contexts prohibitively expensive. To address this, two main categories of plug-and-play solutions have emerged. The first compresses the context into a shorter sequence~\cite{Mu2023LearningTC,Chevalier2023AdaptingLM,Gao2024SelfCPCO,ge2024incontext,li-etal-2025-500xcompressor,li-etal-2023-compressing,DBLP:conf/naacl/TangXLZZHZ25,DBLP:journals/corr/abs-2505-12215,zhao-etal-2025-position,liu2025autoencodingfreecontextcompressionllms}. While effective, these methods are still constrained by the limits of information theory~\cite{shannon1948mathematical}: in general, the compressed length grow with the original. In typical settings, they mainly improve the constant factors in complexity rather than fundamentally changing its asymptotic behavior. The second category utilizes finite-state memory to achieve constant space and linear time~\cite{dai2019transformer,rae2019compressive,bulatov2022recurrent,rodkin2024associativerecurrentmemorytransformer,he-etal-2025-hmt}.
Yet, they struggle to retain fine-grained recent details, and often lack explicit gating mechanisms, making them prone to forgetting critical historical information.

To bridge this gap, we introduce the Collaborative Memory Transformer (CoMeT). As a parameter-efficient and non-invasive memory module, CoMeT is specifically designed to overcome the limitations of prior finite-state models. Its core innovation lies in a synergistic memory system that explicitly addresses both the forgetting of critical information and the loss of recent details. To prevent forgetting, a fixed-size global memory employs a gated update mechanism to distill and shield salient historical information from being overwritten. Concurrently, a temporary memory managed by a First-In-First-Out (FIFO) queue captures fine-grained information from recent chunks, ensuring high-fidelity informational continuity. This design allows CoMeT to elegantly balance the retention of long-term memory with the awareness of long recent context. To enable efficient training on extremely long sequences, we introduce a layer-level pipeline parallelism strategy. This approach yields a $2.7\times$ speedup over the naive context parallel method, making it feasible to fine-tune CoMeT on contexts up to 128k tokens using just $16\times$80GB GPUs.

The capabilities unlocked by CoMeT are substantial. Trained only on 32k-length sequences, CoMeT remarkably extrapolates to accurately retrieve a passkey from any position within a 1M token context (Figure~\ref{sub1}). This feat is achieved with a 21$\times$ inference speedup and a 10$\times$ smaller memory footprint compared to a full-attention baseline at that length. Beyond synthetic tasks, we conduct comprehensive evaluations of CoMeT on both academic language sequence processing tasks and real-world application scenarios. The experimental results demonstrate that CoMeT achieves the highest average score among the efficient methods we evaluated. Notably, on summarization tasks requiring comprehensive understanding, a CoMeT-enhanced model with a memory of just $\sim$2.5k tokens performs on par with a standard Transformer processing the full, uncompressed context.
In summary, CoMeT presents an efficient, practical, and accessible solution to the long-context challenge, pushing the boundaries of what is possible for LLMs.

\section{Related Work}

The pursuit of efficient long-context modeling has evolved along three dominant paradigms: augmenting the standard Transformer with recurrence, developing novel recurrent architectures to replace attention, and compressing context into a more manageable size~\cite{tay2022efficienttransformerssurvey,xiao2023introductiontransformersnlpperspective}. CoMeT operates within the first paradigm, offers a practical alternative to the second, and fundamentally differs from the third in its complexity guarantees.

\paragraph{Recurrent Transformers.}
The chunk-level recurrence is initially introduced into Transformer by Transformer-XL~\cite{dai2019transformer}, which caches hidden states from previous chunks to extend the model's receptive field. Building on this foundation, subsequent work has explored various enhancements. Some methods, like ERNIE-Doc~\cite{ding2021ernie}, concatenate hidden states output at the same layer to grant the model a theoretical receptive field over all preceding content. Compressive Transformer~\cite{rae2019compressive} introduces a dual-queue mechanism to store a compressed representation of older states instead of discarding them. Others, such as RMT~\cite{bulatov2022recurrent} and Memformer~\cite{wu2022memformer}, use memory tokens to recurrently encode historical information chunk by chunk. More recent approaches have designed sophisticated memory structures, such as the associative memory in ARMT~\cite{rodkin2024associativerecurrentmemorytransformer} and the hierarchical system in HMT~\cite{he-etal-2025-hmt}. While these methods successfully achieve $\mathcal{O}(N)$ time and $\mathcal{O}(1)$ space complexity, they suffer from two key limitations that CoMeT addresses. First, many lack explicit gating mechanisms to protect important long-term memories from being overwritten by newer information. Second, they often treat all historical information uniformly, failing to preserve a high-fidelity, fine-grained record of recent events, which is crucial for tasks requiring immediate contextual awareness.

\paragraph{Recurrent Sequence Models.} Another line of work is based on classic recurrent architectures, mainly including Linear Attention mechanisms and State Space Models (SSMs). Linear Attention~\cite{katharopoulos2020transformers} compresses historical key-value information into fixed-size states by removing exponential operations in attention and utilizing the associative property of matrix multiplication; S4~\cite{gu2022efficiently}, S5~\cite{smith2023simplified}, LRU~\cite{pmlr-v202-orvieto23a}, RWKV4/5~\cite{peng-etal-2023-rwkv}, and RetNet~\cite{sun2023retentive} employ data-independent decay mechanisms, while recent advances such as HGRN1/2~\cite{NEURIPS2023_694be354, qin2024hgrn2}, Mamba1/2~\cite{gu2024mamba,pmlr-v235-dao24a}, RWKV6~\cite{peng2024eaglefinchrwkvmatrixvalued}, and GSA~\cite{Zhang2024GatedSA} introduce data-dependent decay mechanisms. DeltaNet~\cite{yang2024parallelizing} and Gated DeltaNet~\cite{yang2024gated} incorporate test-time training to enhance long-term memory capabilities. However, these recurrent sequence methods are specifically designed as architectural alternatives to Transformers and cannot be directly applied to existing pre-trained LLMs in a plug-and-play manner, requiring models to be trained from scratch and thus limiting their adoption in the current LLM ecosystem.

\paragraph{Context Compression.}
Compression methods aim to compress contexts into shorter sequences. Methods such as SelectiveContext~\cite{li-etal-2023-compressing}, LLMLingua~\cite{jiang-etal-2023-llmlingua,pan-etal-2024-llmlingua}, LongLLMLingua~\cite{jiang-etal-2024-longllmlingua}, and EXIT~\cite{hwang2025exitcontextawareextractivecompression} shorten contexts by removing unnecessary portions, while Nano-Capsulator~\cite{chuang-etal-2024-learning}, CompAct~\cite{yoon2024compactcompressingretrieveddocuments}, and FAVICOMP~\cite{jung2025familiarityawareevidencecompressionretrievalaugmented} paraphrase contexts into more concise text. Beyond text-level compression, approaches such as GIST~\cite{mu2023learning}, AutoCompressor~\cite{Chevalier2023AdaptingLM}, LLoCO~\cite{tan-etal-2024-lloco}, ICAE~\cite{ge2024incontext}, 500xCompressor~\cite{li-etal-2025-500xcompressor}, and Activation Beacon~\cite{zhang2025long} compress contexts into shorter compressed embeddings or KV caches. However, under a fixed compression ratio, the length of the compressed sequence still grows linearly with the original context length. This fails to fundamentally alter the asymptotic order of spatiotemporal complexity and can only improve efficiency by reducing constant factors.

\section{Method}
\label{sec:comet_method}

In this section, we introduce the architecture and mechanisms of the Collaborative Memory Transformer (CoMeT). For clarity, Table~\ref{tab:notation} summarizes the key notations used to describe our model. We will first delineate the overall framework in Section~\ref{sec:overall_framework}, then provide a detailed exposition of the global and temporary memory mechanisms in Section~\ref{sec:memory_mechanisms}, and finally, present our layer-level pipeline parallelism strategy for efficient distributed training in Section~\ref{sec:pp}.

\begin{table}[t]
\centering
\scriptsize
\renewcommand{\arraystretch}{1.1} 
\setlength{\tabcolsep}{12pt}
\begin{tabular}{ll}
\toprule
\textbf{Notation} & \textbf{Meaning} \\
\midrule
\rowcolor{gray!10} $\tau$ & Index of the current input chunk. \\
$i$ & Index of the current Transformer layer. \\
\rowcolor{gray!10} $\mathbf{H}^i_{\tau}$ & Hidden states of the $\tau$-th chunk at layer $i$. \\
$\mathbf{G}^i_{\tau}$ & Global memory tokens for chunk $\tau$ at layer $i$. \\
\rowcolor{gray!10} $\mathbf{T}^i_{\tau}$ & Temporary memory tokens for chunk $\tau$ at layer $i$. \\
$\mathbf{S}^i_{\tau}$ & Persistent global state for chunk $\tau$ at layer $i$. \\
\rowcolor{gray!10} $\mathbf{C}^i_{\tau}$ & Compression tokens for chunk $\tau$ at layer $i$. \\
$\mathbf{R}^i_{\tau}$ & Readout tokens for chunk $\tau$ at layer $i$. \\
\rowcolor{gray!10} $m$ & Number of readout tokens. \\
$\mathrm{TL}(\cdot)$ & A single Transformer layer computation. \\
\rowcolor{gray!10} $\mathrm{RLA}(\cdot)$ & Residual Low-Rank Adapter module. \\
$d_{\text{model}}$ & Hidden dimension of the model. \\
\rowcolor{gray!10} $r$ & Rank of the low-rank projection in the RLA. \\
\bottomrule
\end{tabular}
\caption{Notation used to describe the CoMeT architecture in Section~\ref{sec:comet_method} and Figure~\ref{fig:TransformerLayer}.\label{tab:notation}}
\end{table}

\begin{figure*}[htbp]
    \centering
    \includegraphics[width=1\linewidth]{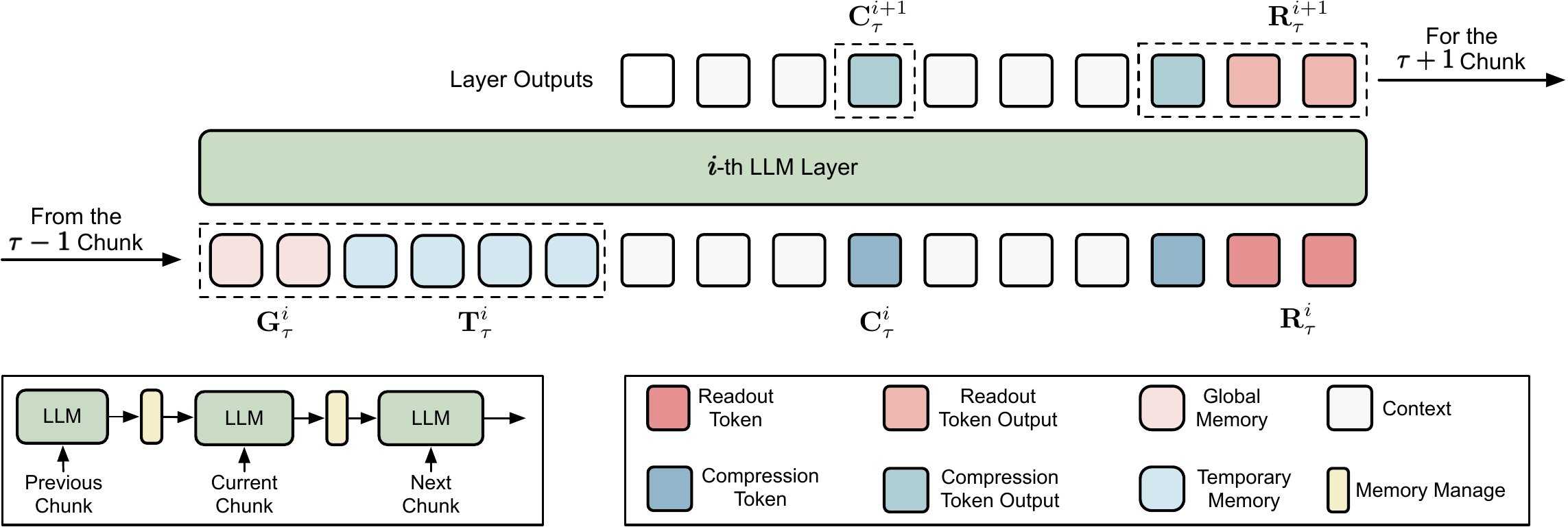}
    \caption{Overview of CoMeT. At layer $i$, the global memory $\mathbf{G}^i_{\tau}$ and temporary memory $\mathbf{T}^i_{\tau}$ are prepended to the current chunk's hidden states $\mathbf{H}^i_{\tau}$. Compression tokens $\mathbf{C}^i_{\tau}$ are interleaved within the hidden states for fine-grained information capture, while readout tokens $\mathbf{R}^i_{\tau}$ are appended at the end to distill key information for updating the global state. All tokens interact through causal self-attention, enabling the model to retrieve relevant historical information while processing the current chunk.}
    \label{fig:TransformerLayer}
\end{figure*}

\subsection{Overall Framework}
\label{sec:overall_framework}
Following prior work, CoMeT processes the input context in a chunk-by-chunk manner. As illustrated in Figure~\ref{fig:TransformerLayer}, at the $i$-th Transformer layer, when processing the $\tau$-th input chunk, the model prepends the global memory $\mathbf{G}^i_{\tau}$ and temporary memory $\mathbf{T}^i_{\tau}$ to the chunk's hidden states $\mathbf{H}^i_{\tau}$. Through the causal self-attention mechanism, $\mathbf{H}^i_{\tau}$ can retrieve relevant information from both memories to inform next-token prediction. Concurrently, we interleave a set of compression tokens $\mathbf{C}^i_{\tau}$ within $\mathbf{H}^i_{\tau}$ to distill fine-grained local information. Finally, $m$ readout tokens $\mathbf{R}^i_{\tau}$ are appended to the sequence to summarize the chunk's most salient content. The overall computation of a single Transformer layer is thus formulated as: $\mathbf{H}^{i+1}_{\tau}, \mathbf{C}^{i+1}_{\tau}, \mathbf{R}^{i+1}_{\tau} = \mathrm{TL}(\mathbf{G}^i_{\tau}, \mathbf{T}^i_{\tau}, \mathbf{H}^{i}_{\tau}, \mathbf{C}^{i}_{\tau}, \mathbf{R}^{i}_{\tau})$, where $\mathrm{TL}$ denotes the Transformer layer computation.


\subsection{Collaborative Memory Mechanisms}
\label{sec:memory_mechanisms}

CoMeT's memory system is composed of two synergistic components: a global memory for long-range dependencies and a temporary memory for recent context.

\paragraph{Global Memory.}
\begin{figure}[htbp]
    \centering
    \includegraphics[width=0.99\linewidth]{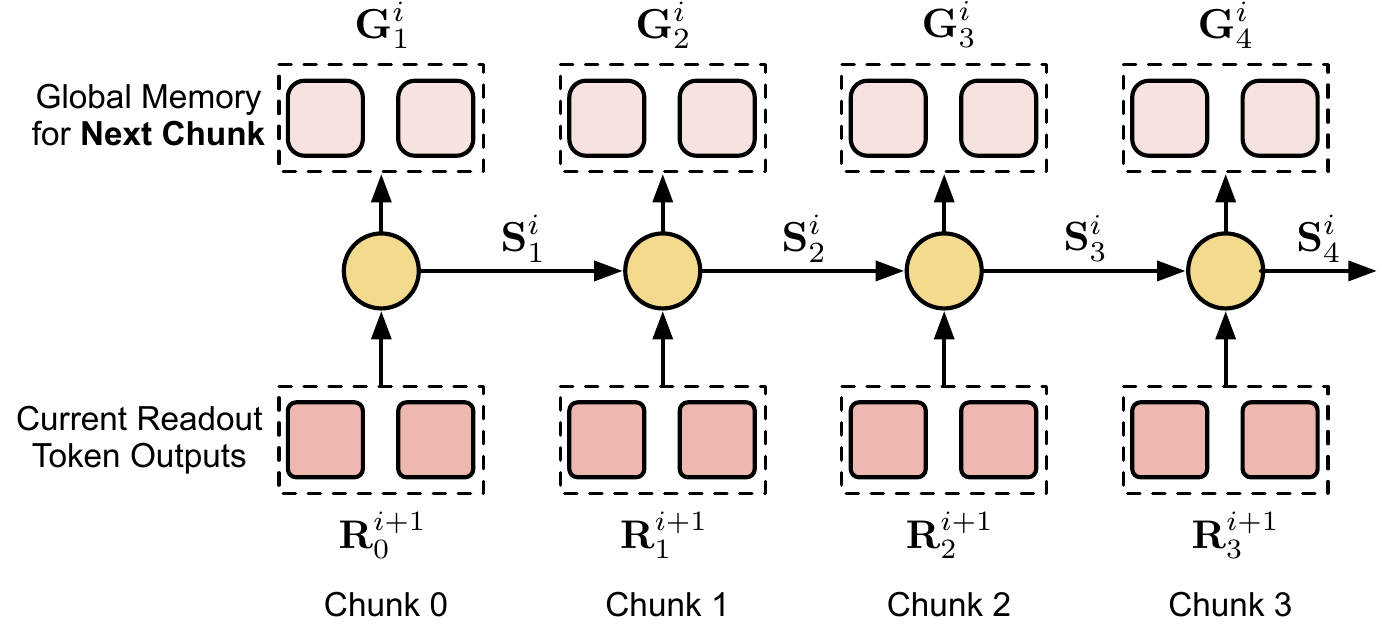}
    \caption{Architecture of the global memory mechanism. At each layer $i$ and chunk $\tau$, the global state $\mathbf{S}^i_{\tau}$ is transformed by a $\mathrm{RLA}$ to produce the global memory $\mathbf{G}^i_{\tau}$. The state is then updated for the next chunk via a gating mechanism that selectively integrates information from the normalized readout tokens $\mathbf{R}^{i+1}_{\tau}$.}
    \label{fig:GlobalMemory}
\end{figure}
As depicted in Figure~\ref{fig:GlobalMemory}, the global memory $\mathbf{G}^i_{\tau}$ is derived 
from a persistent global state $\mathbf{S}^i_{\tau}$. 
Our preliminary experiments reveal that introducing an excessive number of parameters for 
this state-to-memory transformation degrades performance. 
We therefore employ a parameter-efficient module we term the 
\textbf{Residual Low-Rank Adapter (RLA)}, which transforms a state vector by adding a low-rank projection:
\begin{equation}
    \mathrm{RLA}(\mathbf{X}) =\mathbf{X} + \mathbf{W}_{\text{up}} ( \mathbf{W}_\text{down} \mathbf{X})
    \label{eq:rla}
\end{equation}
where the projection matrices are $\mathbf{W}_{\text{up}} \in \mathbb{R}^{d_{\text{model}} \times r}$ and $\mathbf{W}_\text{down} \in \mathbb{R}^{r \times d_{\text{model}}}$. The global memory is thus computed as $\mathbf{G}^i_{\tau} = \mathrm{RLA}(\mathbf{S}^i_{\tau})$. This additive, low-rank structure ensures minimal parameter overhead while promoting stable training. We set the rank $r=8$ unless stated otherwise\footnote{The CoMeT module adds only 3.95M parameters (0.098\%) to the Qwen3-4B-Instruct-2507 model when rank=8.}.

The global state for the next chunk $\mathbf{S}^i_{\tau+1}$ is updated using the output readout tokens $\mathbf{R}^{i+1}_{\tau}$. Prior to the update, these tokens are normalized to form a candidate state: $\tilde{\mathbf{S}}^i_{\tau+1} = \mathrm{RMSNorm}(\mathbf{R}^{i+1}_{\tau})$. We then employ a gating mechanism for the update:
\begin{equation}
    \mathbf{S}^i_{\tau+1} = \mathbf{g} \odot \mathbf{S}^i_{\tau} + (\mathbf{1}-\mathbf{g}) \odot \tilde{\mathbf{S}}^i_{\tau+1}
    \label{eq:global_update}
\end{equation}
where the gate $\mathbf{g} = \sigma(\mathbf{W}_\text{g}([\mathbf{S}^i_{\tau}; \tilde{\mathbf{S}}^i_{\tau+1}]))$. Here, $[\cdot; \cdot]$ denotes concatenation along the feature dimension, $\mathbf{W}_\text{g} \in \mathbb{R}^{2d_{\text{model}} \times 1}$ is a learnable weight matrix, and $\sigma$ represents the sigmoid function. This mechanism allows the state to selectively absorb new information while shielding essential historical information from being overwritten. Furthermore, this additive update structure, reminiscent of gates in LSTMs and GRUs, creates a more direct path for gradient flow across chunks.

\paragraph{Temporary Memory.}
\begin{figure}[htbp]
    \centering
    \includegraphics[width=0.99\linewidth]{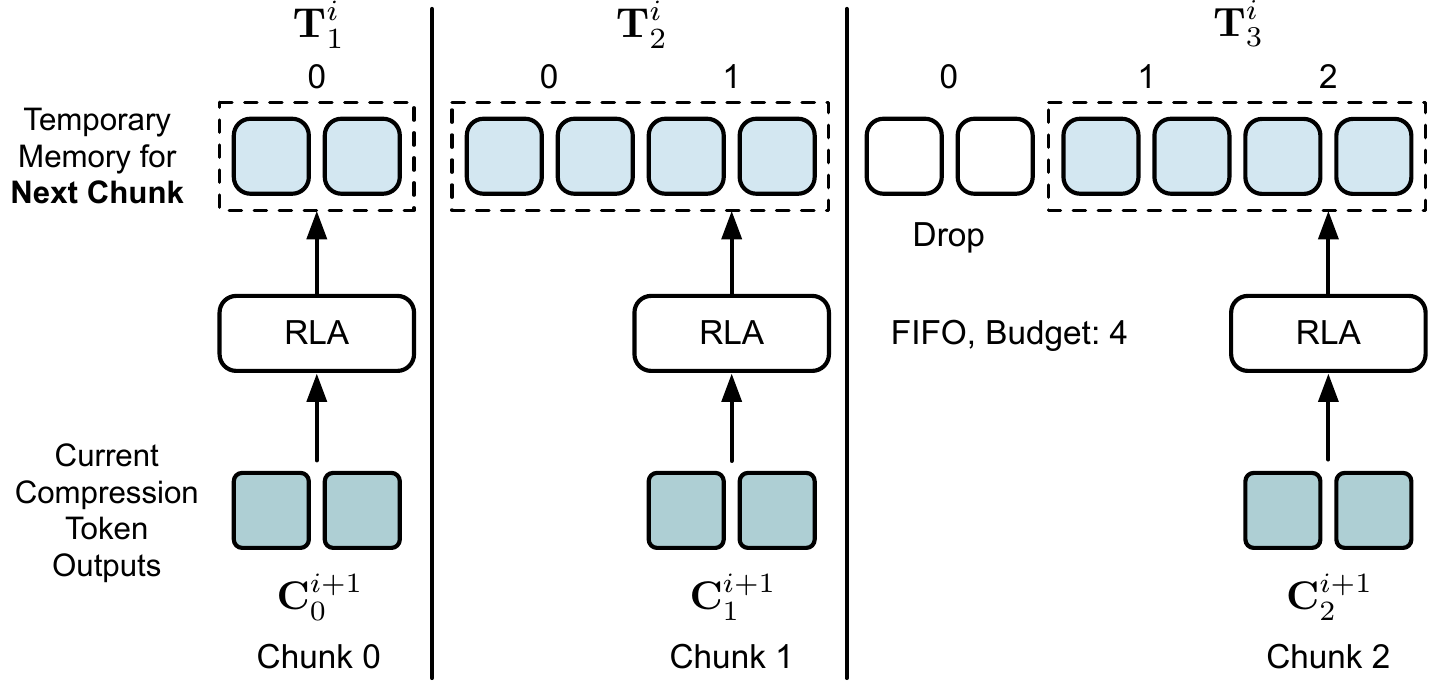}
    \caption{The architecture of the temporary memory mechanism. \textbf{CoMeT} employs a fixed-capacity FIFO queue to manage compressed representations of recent chunks. As new information from the current chunk is enqueued, the oldest memory entry is discarded. This rolling window of memory provides the model with a high-resolution view of the most recent context while maintaining a constant memory footprint.}
    \label{fig:TemporaryMemory}
\end{figure}
As shown in Figure~\ref{fig:TemporaryMemory}, we manage the temporary memory $\mathbf{T}^i_{\tau}$ using a First-In-First-Out (FIFO) queue of fixed capacity. New memory entries are derived from the output compression tokens $\mathbf{C}^{i+1}_{\tau}$. These tokens are first processed by $\mathrm{RMSNorm}$ and then transformed using the same $\mathrm{RLA}$ module (as defined in Eq.~\ref{eq:rla}) before being enqueued into the FIFO queue.

The FIFO nature of the queue preserves the temporal continuity of information from recent chunks. As a new entry is added, the oldest is discarded. This mechanism, combined with fine-grained compression, allows the model to maintain a high-resolution memory of the immediate context. From an optimization perspective, the FIFO queue also creates direct gradient paths back to recent chunks held in memory, enhancing training stability.

\subsection{Efficient Long Context Training}
\label{sec:pp}

\begin{figure}[htbp]
    \centering
    \includegraphics[width=1\linewidth]{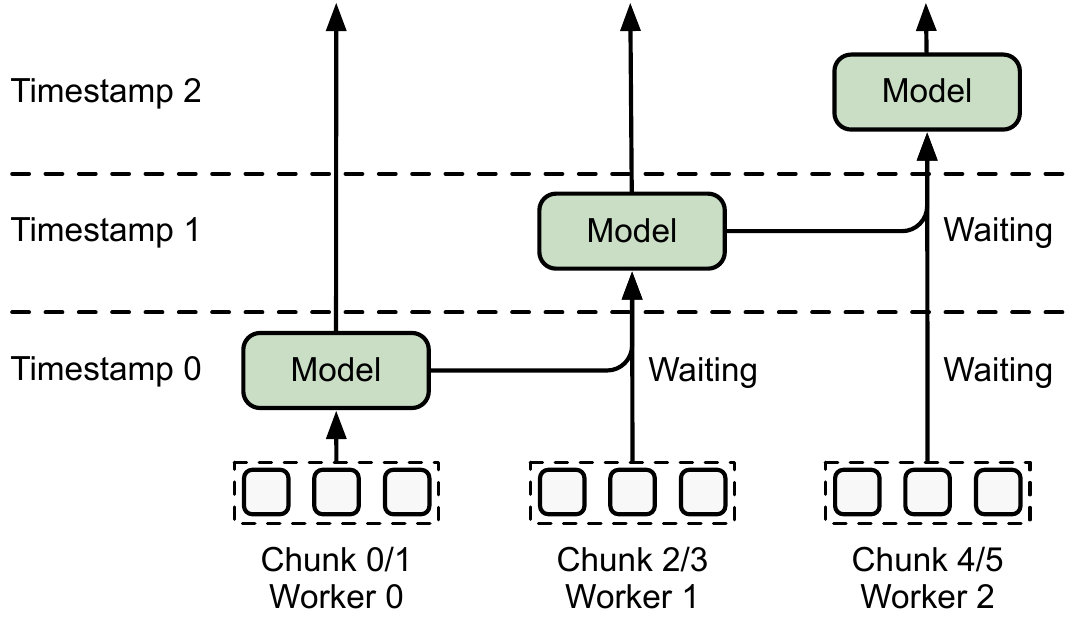}
    \caption{Naive context parallelism. Workers process chunks sequentially. Worker $j+1$ must wait for worker $j$ to complete its entire computation before starting, creating a large pipeline bubble and leading to significant resource under-utilization.}
    \label{fig:ppv1}
\end{figure}
Training CoMeT on extremely long sequences necessitates a distributed approach. A naive context parallelism strategy, as depicted in Figure~\ref{fig:ppv1}, distributes chunks across GPU workers, 
with memory states passed between them via P2P communication. 
This method, however, suffers from a strict serial dependency, 
as each worker must wait for the previous one to complete its entire forward pass. 
This creates a large pipeline bubble, leaving most workers idle and leading to severe under-utilization of computational resources.

\begin{figure}[htbp]
    \centering
    \includegraphics[width=1\linewidth]{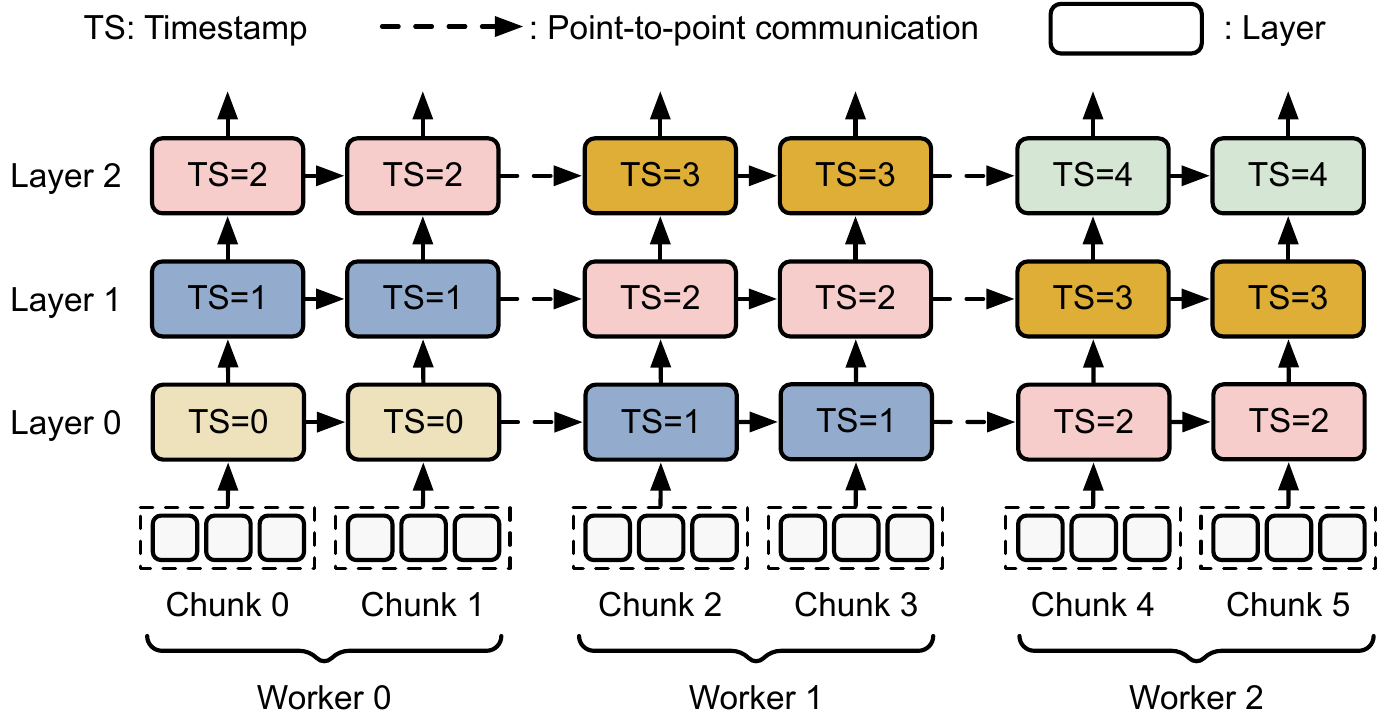}
    \caption{Our proposed layer-level pipeline parallelism. Computation and communication are interleaved at the layer level. Worker $j+1$ begins processing layer $i$ as soon as it receives the necessary state from worker $j$, significantly reducing the pipeline bubble and maximizing hardware utilization.}
    \label{fig:ppv2}
\end{figure}

To address this inefficiency, we propose a fine-grained pipeline parallelism method that 
interleaves computation and communication at the layer level (Figure~\ref{fig:ppv2}). 
Rather than waiting for a full chunk computation, a worker, upon finishing layer $i$, 
immediately transmits the required memory state to the next worker. 
This enables the receiving worker to start on layer $i$ while the sending worker 
concurrently advances to layer $i+1$. 
By maximizing worker concurrency, this strategy dramatically reduces idle time, 
boosts training throughput, and enables efficient scaling to very long sequences.
This approach makes it feasible to train a Qwen3-4B-based \textbf{CoMeT} model with a 128K context length using just 16$\times$ 80GB GPUs. The visualization of pipeline bubbles for Layer-Level Pipeline Parallelism strategies is provided in Appendix~\ref{appendix:bubble}.

\begin{table*}[htbp]
\setlength{\tabcolsep}{4pt} 
\centering
\small 

\begin{tabular}{l c c c c c c c c c}
\toprule
  & \multirow{2}{*}{\textbf{Memory}} & \textbf{GovRep} & \textbf{SumScr} & \textbf{QMSum} & \textbf{Qspr} & \textbf{Nrtv} & \textbf{QALT} & \textbf{CNLI} & \multirow{2}{*}{\textbf{Avg}} \\

\cmidrule(lr){3-3} \cmidrule(lr){4-4} \cmidrule(lr){5-5} \cmidrule(lr){6-6} \cmidrule(lr){7-7} \cmidrule(lr){8-8} \cmidrule(lr){9-9}  

& & R-1/2/L & R-1/2/L & R-1/2/L & F1 & F1 & F1 & EM \\

\midrule 

\specialrule{0em}{1pt}{1pt}
Full Attn & Full Context &52.7/17.1/20.3 & 19.1/4.2/10.2 & 16.3/4.6/10.1 & 3.5 & 2.5 & 3.8 & 0.0 & 7.87 \\
Full Attn (FT) & Full Context & \textbf{61.0}/\textbf{31.9}/\textbf{33.0} & \textbf{32.5}/\textbf{7.6}/\textbf{19.0} & \textbf{37.4}/\textbf{12.9}/\textbf{25.6} & \textbf{40.3} & \textbf{22.1} & \textbf{64.2} & \textbf{89.1} & \textbf{42.23} \\

\cdashline{1-10}
\specialrule{0em}{3pt}{1pt}
\multicolumn{10}{l}{\textit{Compression}} \\
\specialrule{0em}{1pt}{1pt}
LongLLMLingua & 3072 tok& 38.0/14.5/20.0 & 28.2/5.4/16.7 & \textbf{34.6}/\textbf{11.4}/\textbf{23.3} & \textbf{35.7} & 19.2 & \textbf{65.9} & 83.9 & \textbf{37.36} \\
LLMLingua2 & 3072 tok & 32.1/12.5/19.0 & \textbf{29.8}/6.2/\textbf{17.9} & 32.9/9.4/22.0 & 35.4 & 16.4 & 61.1 & \textbf{88.2} & 36.38 \\
EXIT & 3072 tok & 48.6/21.3/24.2 & 28.8/5.8/17.4 & 32.3/8.9/21.4 & 35.4 & 14.9 & 59.9 & 86.5 & 36.94 \\
ICAE & 192×16 tok& 25.4/5.5/17.4 & 21.2/3.3/13.9 & 28.7/7.8/20.6 & 18.5 & 15.7 & 54.9 & 74.2 & 29.04 \\
500xCompressor & 192×16 tok& 34.4/12.4/20.9 & 23.5/4.4/14.9 & 24.1/7.1/18.1 & 23.0 & 19.0 & 56.3 & 82.6 & 32.54 \\
ActivationBeacon & 256×16 tok& \textbf{52.3}/\textbf{25.0}/\textbf{27.5} & 28.0/\textbf{6.5}/17.1 & 31.8/10.2/22.7 & 33.5 & \textbf{23.2} & 56.8 & 25.8 & 30.71 \\

\cdashline{1-10}
\specialrule{0em}{3pt}{1pt}
\multicolumn{10}{l}{\textit{Finite-state}} \\
\specialrule{0em}{1pt}{1pt}

Transformer-XL & ws=5120 & 51.2/23.0/27.0 & 30.7/6.4/17.8 & 27.2/5.7/18.6 & 35.5 & 4.5 & 33.6 & 88.1 & 31.83 \\
SWA & ws=5120 & 55.3/26.9/29.6 & 30.7/6.8/17.9 & 32.4/9.1/21.7 & \textbf{39.1} & 16.1 & 54.8 & \textbf{88.3} & 38.24 \\
HMT & ms=3072 & 47.3/15.0/21.9 & 29.0/3.7/15.9 & 31.9/7.1/20.1 & 16.8 & 11.3 & 53.5 & 77.1 & 30.31 \\
\cellcolor[rgb]{0.925,0.957,1}{CoMeT} & \cellcolor[rgb]{0.925,0.957,1}ms=2560 & \cellcolor[rgb]{0.925,0.957,1}\textbf{62.5}/\textbf{31.1}/\textbf{33.4} & \cellcolor[rgb]{0.925,0.957,1}\textbf{33.4}/\textbf{8.3}/\textbf{19.8} & \cellcolor[rgb]{0.925,0.957,1}\textbf{35.6}/\textbf{12.0}/\textbf{24.6} & \cellcolor[rgb]{0.925,0.957,1}35.5 & \cellcolor[rgb]{0.925,0.957,1}\textbf{22.6} & \cellcolor[rgb]{0.925,0.957,1}\textbf{56.0} & \cellcolor[rgb]{0.925,0.957,1}86.9 & \cellcolor[rgb]{0.925,0.957,1}\textbf{40.10} \\

\midrule 

\specialrule{0em}{1pt}{1pt}
Avg. Length & & 10,535 & 8,617 & 13,291 & 5,462 & 19,250 & 6,085 & 2,210 & \\

\bottomrule 
\end{tabular}
\caption{Results on \scrolls{} benchmark. All efficient methods use $\sim$3k memory budget. CoMeT outperforms other efficient methods and matches the fine-tuned full attention baseline on summarization tasks. GovRep, SumScr, QMSum, Qspr, Nrtv, QALT, and CNLI denote GovReport, SummScreenFD, QMSum, Qasper, NarrativeQA, QuALITY, and ContractNLI, respectively. FT denotes Fine-Tuned. SWA denotes Sliding Window Attention.}
\label{tab:scrolls_results_detailed}
\end{table*}

\section{Experiments}

To comprehensively evaluate CoMeT, we conduct experiments across three dimensions: (1) academic benchmarks to assess fundamental long-context language understanding capabilities, (2) real-world scenarios to validate practical applicability, and (3) passkey retrieval tasks to examine information extraction in extremely long contexts.

\begin{table}[htbp]
    \setlength{\tabcolsep}{10.5pt} 
    \centering
    \small
    \begin{tabular}{l cc cc}
        \toprule
        \multirow{2}{*}{ } & \multicolumn{2}{c}{\textbf{2WikiMQA}} & \multicolumn{2}{c}{\textbf{HotpotQA}} \\
        \cmidrule(lr){2-3} \cmidrule(lr){4-5}
        & \textbf{EM} & \textbf{F1} & \textbf{EM} & \textbf{F1} \\
        \midrule
        Full Attn  & 75.4 & 80.8 & 65.0 & 78.9 \\
        \cellcolor[rgb]{0.925,0.957,1}CoMeT & \cellcolor[rgb]{0.925,0.957,1}\textbf{75.5} & \cellcolor[rgb]{0.925,0.957,1}\textbf{81.0} & \cellcolor[rgb]{0.925,0.957,1}\textbf{65.9} & \cellcolor[rgb]{0.925,0.957,1}\textbf{80.0} \\
        \midrule
        Avg. Length & \multicolumn{2}{c}{1033} & \multicolumn{2}{c}{1443} \\
        \bottomrule
    \end{tabular}
    \caption{Performance comparison on 2WikiMQA and HotpotQA. The last row shows the average context length for each dataset's development set.}
    \label{tab:qa_results_comparison_with_length}
\end{table}

\subsection{Baseline Methods}

We benchmark CoMeT against various plug-and-play methods, including context compression (e.g., LongLLMLingua, Activation Beacon) and finite-state models (e.g., Transformer-XL, Sliding Window Attention). Full Attention serves as the performance upper bound.

\subsection{Experimental Setup}

By default, our experiments employ Qwen3-4B-Instruct-2507~\cite{qwen3technicalreport} as the backbone model. To ensure a fair comparison, all efficient methods are allocated a comparable memory budget of approximately 3k tokens. We fine-tune relevant models for 3 epochs on a 32k context length using a unified training configuration. Detailed parameters for each baseline and our training setup, including learning rates and optimizer settings, are provided in Appendix~\ref{sec:setup}.

\subsection{Evaluation Results}


\paragraph{Language Sequence Processing Tasks.}
We evaluate CoMeT on the \scrolls{} benchmark~\cite{shaham-etal-2022-scrolls}, which includes GovReport~\cite{huang-etal-2021-efficient}, SummScreenFD~\cite{chen-etal-2022-summscreen}, QMSum~\cite{zhong-etal-2021-qmsum}, Qasper~\cite{dasigi-etal-2021-dataset}, NarrativeQA~\cite{10.1162/tacl_a_00023}, QuALITY~\cite{pang-etal-2022-quality}, and ContractNLI~\cite{koreeda-manning-2021-contractnli-dataset}. To assess performance on shorter sequences, we also include 2WikiMQA~\cite{ho-etal-2020-constructing} and HotpotQA~\cite{yang-etal-2018-hotpotqa}. All fine-tunable models are trained for 3 epochs on a mixed dataset with up to 32k context length. Further details about these tasks are provided in Appendix~\ref{app:dataset_details}.

As shown in Table~\ref{tab:scrolls_results_detailed}, CoMeT achieves the highest average score among all efficient methods. Crucially, on summarization tasks that require a holistic understanding of the input (GovRep, SumScr), CoMeT performs on par with the fine-tuned Full Attention baseline. Beyond the standard benchmark splits, we further evaluate CoMeT's extrapolation ability on long-context sequences exceeding its training length (Appendix~\ref{app:extrapolation}), where it continues to outperform SWA by a large margin. On shorter sequences (Table~\ref{tab:qa_results_comparison_with_length}), CoMeT naturally matches Full Attention performance, as the entire input fits within a single chunk.

\begin{table}[htbp]
    \setlength{\tabcolsep}{4pt} 
    \centering
    \small
    \begin{tabular}{l l c c}
        \toprule
        & \textbf{Memory} & \textbf{UQA} & \textbf{Terminal Bench} \\
        \midrule
        Full Attn & 4k  & 51.3 & -- \\
        Full Attn & 32k  & \textbf{81.3} & -- \\
        Full Attn & 128k & -- & \textbf{21.33} \\
        \midrule
        xRAG & -- & 76.0 & -- \\
        \cellcolor[rgb]{0.925,0.957,1}CoMeT & \cellcolor[rgb]{0.925,0.957,1}4k  & \cellcolor[rgb]{0.925,0.957,1}\textbf{78.7} & \cellcolor[rgb]{0.925,0.957,1}-- \\
        \cellcolor[rgb]{0.925,0.957,1}CoMeT & \cellcolor[rgb]{0.925,0.957,1}5k & \cellcolor[rgb]{0.925,0.957,1}-- & \cellcolor[rgb]{0.925,0.957,1}20.27 \\
        \bottomrule
    \end{tabular}
    \caption{Real-world application results on user behavior sequence QA and agent tasks. For user behavior QA, CoMeT outperforms the xRAG baseline and significantly improves over 4k truncated Full Attention. For code tasks, experiments are conducted at 128k sequence length with the Qwen3-8B model (8B instead of 4B, as the 4B model lacks the fundamental capabilities to solve the complex tasks in Terminal-Bench), where Full Attention training is enabled via Megatron-LM's~\cite{megatron-lm} sequence parallelism. CoMeT uses chunk size 4096 and memory size 1024 (G) + 4096 (T).}
    \label{tab:real_world_results}
\end{table}

\paragraph{Real-World Application Scenarios.}

To demonstrate CoMeT's real-world utility, we evaluate it on two application-driven benchmarks: User Behavior QA (UQA) and a long-context agent task. Details are in Appendix \ref{app:dataset_details}.
The UQA benchmark requires reasoning over thousands of user interactions. On a real-world e-commerce dataset, CoMeT outperforms a strong industry xRAG\footnote{xRAG encodes each behavior item in the user sequence into a single vector and concatenates these vectors with the question as input to the decoder.} baseline by 2.7 accuracy points and a naive 4k Truncation baseline by 27.4 points (Table~\ref{tab:real_world_results}).
For the agent task, we use iflow-cli\footnote{\url{https://github.com/iflow-ai/iflow-cli}} as the agent framework, fine-tune the model using 128k-token trajectories, and report results on Terminal-Bench~\cite{tbench_2025}. This extreme context length precludes training other efficient methods. Benefiting from our layer-level pipeline parallelism, CoMeT's training is $2.7\times$ faster than naive context parallelism. It achieves performance competitive with a full-attention model while being vastly more efficient, validating CoMeT as a practical solution for deploying LLMs in real-world environments.


\paragraph{Passkey Retrieval Task.}
To evaluate CoMeT's performance in extreme-length contexts, we use a passkey retrieval task requiring finding a \textbf{5- to 7-digit} passkey within distractor text (details in Appendix~\ref{sec:passkey}). After fine-tuning for 1500 steps on 32k-length sequences, CoMeT demonstrates remarkable extrapolation, successfully retrieving the passkey from any position within a 1M-token context (Figure~\ref{sub1}).

\begin{figure}
    \centering
    \includegraphics[width=1\linewidth]{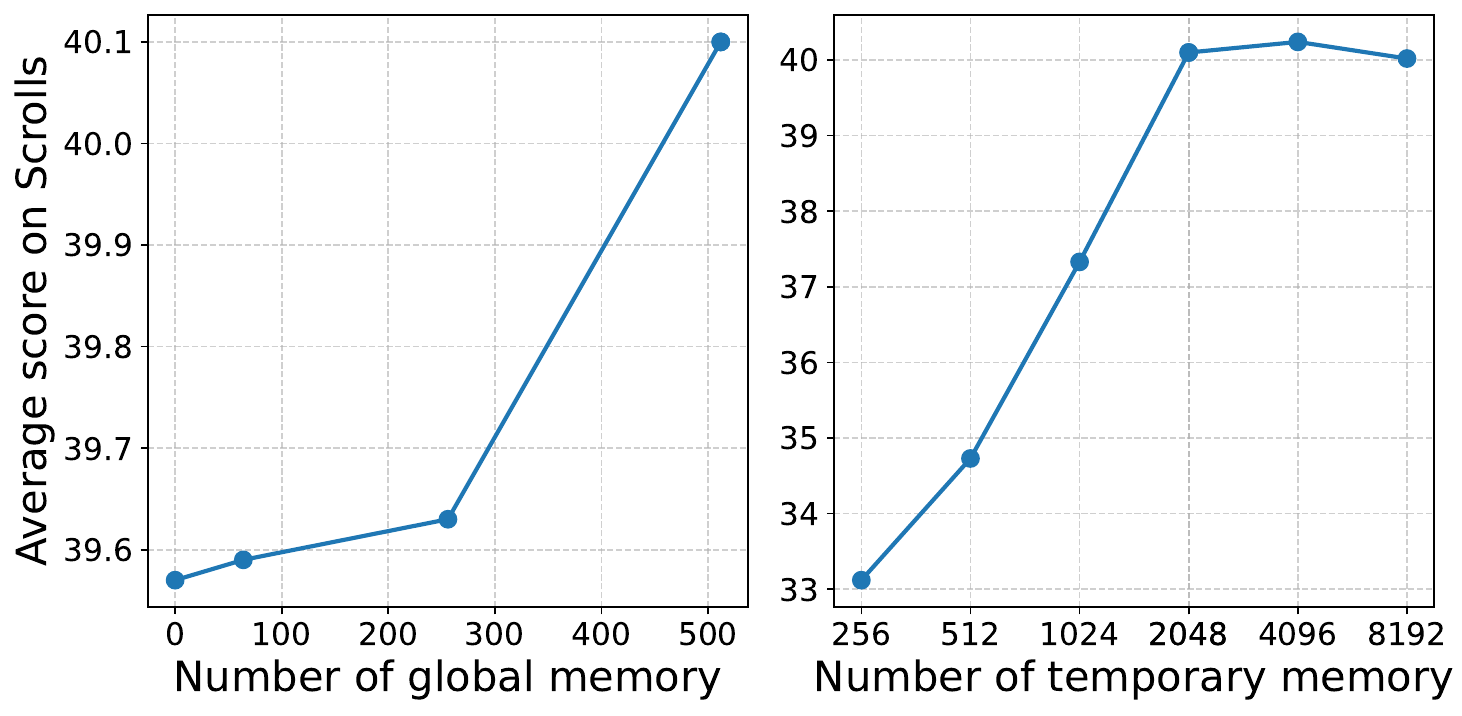}
        \caption{Performance impact of varying global and temporary memory sizes on the \scrolls{} benchmark.}
    \label{fig:memory_allo}
\end{figure}

\begin{figure}[t]
\centering
\captionsetup[sub]{font=small, labelfont=small}
 \begin{minipage}{0.5\textwidth}  
        \centering
        \begin{minipage}{0.48\textwidth}
            \centering
            \subcaptionbox{Global 1024, Temp 2048\label{fig:g1024_t2048}}{
            \includegraphics[width=\textwidth]{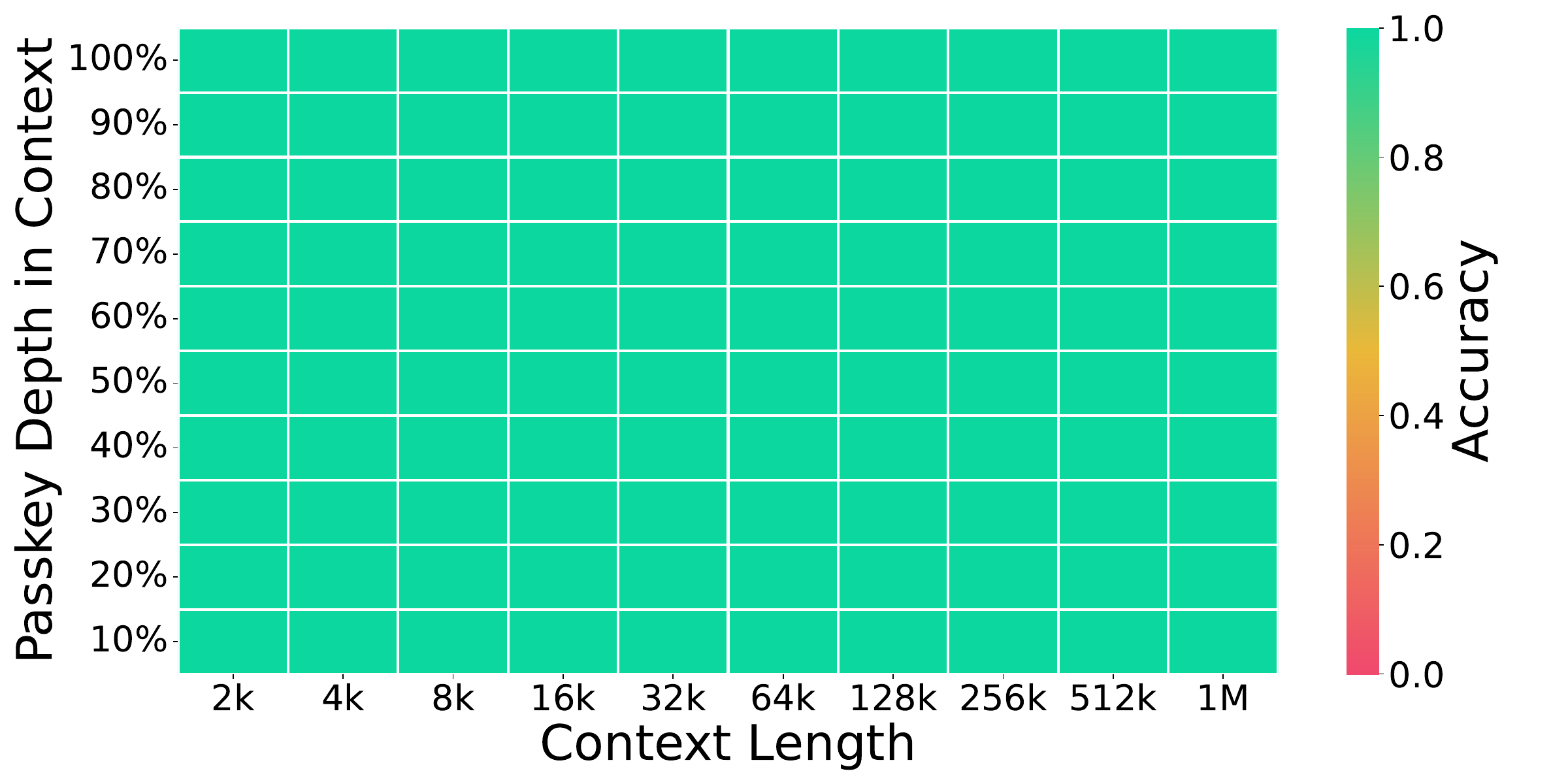}
            }
        \end{minipage}
        \hfill
        \begin{minipage}{0.48\textwidth}
            \centering
            \subcaptionbox{Global 3072, Temp 0\label{fig:g3072_t0}}{
            \includegraphics[width=\textwidth]{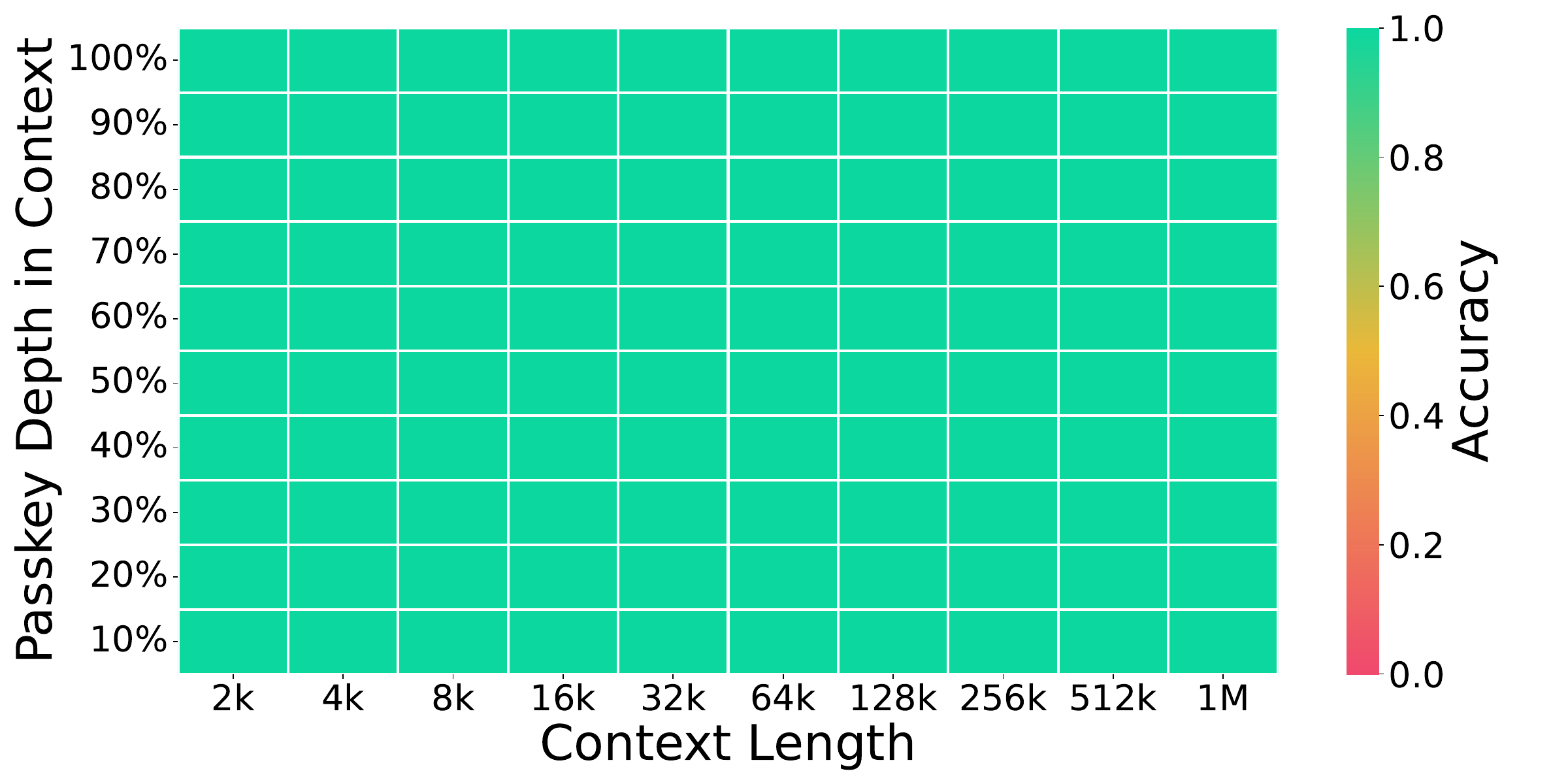}
            }
        \end{minipage}
        
        \vspace{0.1cm}
        
        \begin{minipage}{0.48\textwidth}
            \centering
            \subcaptionbox{Global 8, Temp 3072\label{fig:g8_t3072}}{
             \includegraphics[width=\textwidth]{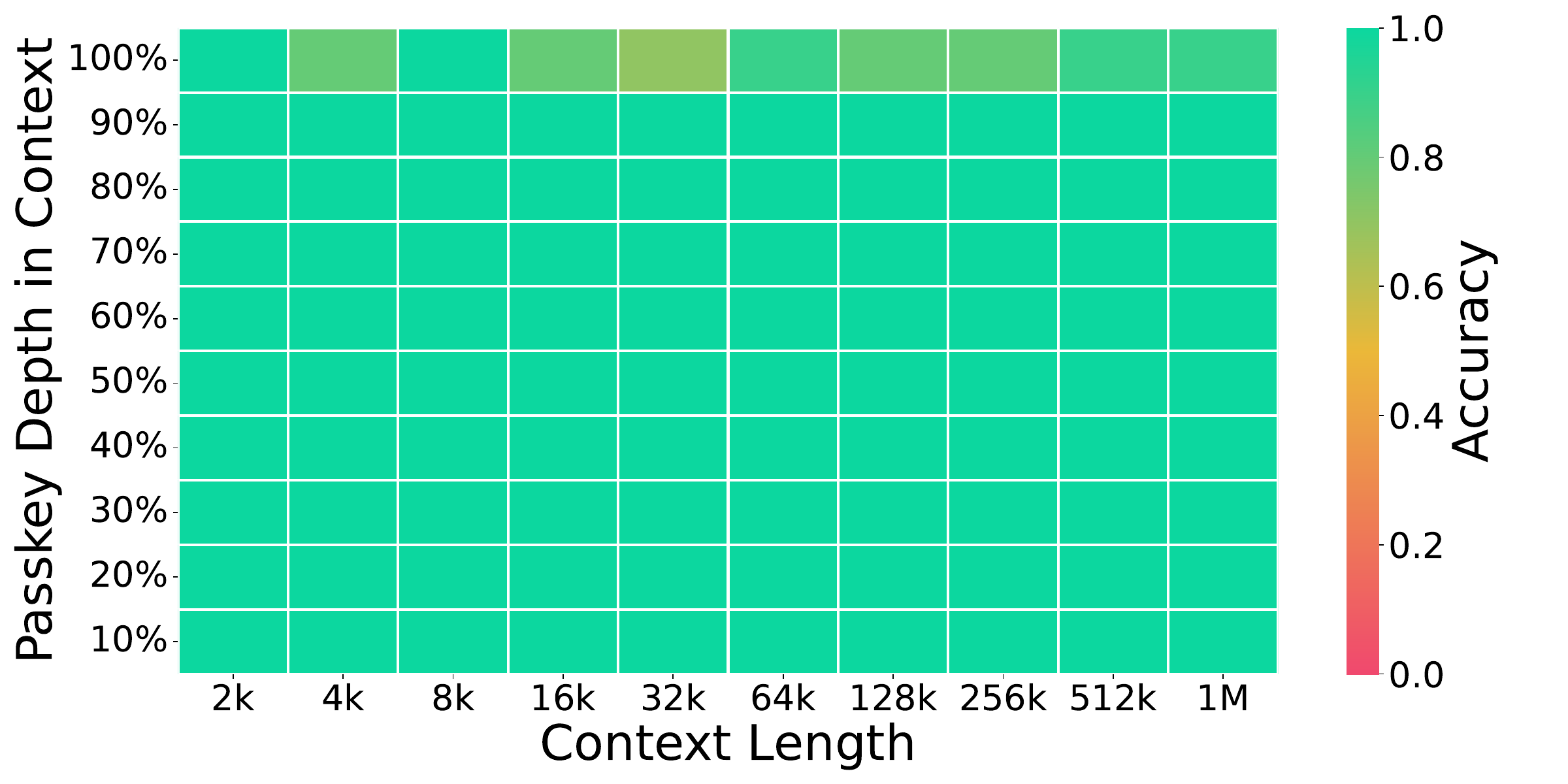}
            }
        \end{minipage}
        \hfill
        \begin{minipage}{0.48\textwidth}
            \centering
            \subcaptionbox{Global 3072, no gate\label{fig:g3072_b0_without_gate}}{
             \includegraphics[width=\textwidth]{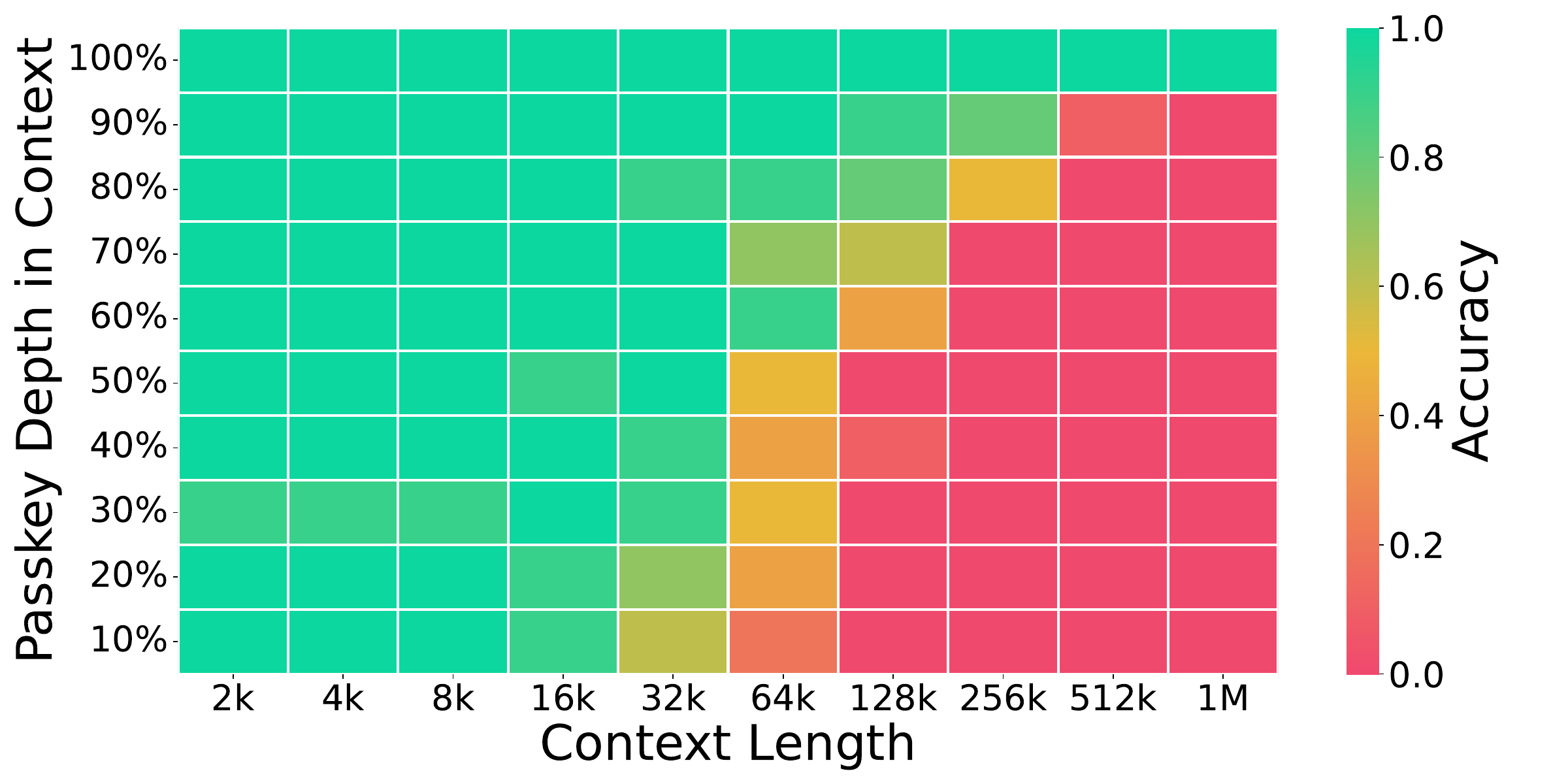}
            }
        \end{minipage}
 \end{minipage}
\caption{Passkey retrieval accuracy under different memory configurations. (a) Balanced configuration with 1024 global and 2048 temporary memory. (b) Global-only configuration using all 3072 tokens for global memory. (c) Temporary-only configuration with minimal (8) global memory. (d) Global memory without gating mechanism, demonstrating the critical role of gates in long-term information retention.}
\label{fig:MTP_Fig}
\end{figure}

\section{Analysis}

\begin{table*}[htbp]
\setlength{\tabcolsep}{3.3pt}
\centering
\small
\begin{tabular}{l l c c c c c c c c}
\toprule
& \multirow{2}{*}{\textbf{Memory}} & \textbf{GovRep} & \textbf{SumScr} & \textbf{QMSum} & \textbf{Qspr} & \textbf{Nrtv} & \textbf{QALT} & \textbf{CNLI} & \multirow{2}{*}{\textbf{Avg}} \\
\cmidrule(lr){3-3} \cmidrule(lr){4-4} \cmidrule(lr){5-5} \cmidrule(lr){6-6} \cmidrule(lr){7-7} \cmidrule(lr){8-8} \cmidrule(lr){9-9}
& & R-1/2/L & R-1/2/L & R-1/2/L & F1 & F1 & F1 & EM \\
\midrule
\specialrule{0em}{1pt}{1pt}
\multicolumn{10}{l}{\textit{Qwen3-4B-Instruct-2507}} \\
\specialrule{0em}{1pt}{1pt}
Full Attn (FT) & Full Context & 61.0/\textbf{31.9}/33.0 & 32.5/7.6/19.0 & \textbf{37.4}/\textbf{12.9}/\textbf{25.6} & \textbf{40.3} & 22.1 & \textbf{64.2} & \textbf{89.1} & \textbf{42.23} (100\%) \\
SWA & ws=5120 & 55.3/26.9/29.6 & 30.7/6.8/17.9 & 32.4/9.1/21.7 & 39.1 & 16.1 & 54.8 & 88.3 & 38.24 (90.6\%) \\
\cellcolor[rgb]{0.925,0.957,1}CoMeT & \cellcolor[rgb]{0.925,0.957,1}ms=2560 & \cellcolor[rgb]{0.925,0.957,1}\textbf{62.5}/31.1/\textbf{33.4} & \cellcolor[rgb]{0.925,0.957,1}\textbf{33.4}/\textbf{8.3}/\textbf{19.8} & \cellcolor[rgb]{0.925,0.957,1}35.6/12.0/24.6 & \cellcolor[rgb]{0.925,0.957,1}35.5 & \cellcolor[rgb]{0.925,0.957,1}\textbf{22.6} & \cellcolor[rgb]{0.925,0.957,1}56.0 & \cellcolor[rgb]{0.925,0.957,1}86.9 & \cellcolor[rgb]{0.925,0.957,1}40.10 (95.0\%) \\
\cdashline{1-10}
\specialrule{0em}{3pt}{1pt}
\multicolumn{10}{l}{\textit{Qwen3-14B}} \\
\specialrule{0em}{1pt}{1pt}
Full Attn (FT) & Full Context & 62.1/\textbf{32.9}/34.2 & 35.1/8.6/20.5 & \textbf{37.7}/\textbf{13.3}/\textbf{25.5} & 41.7 & 24.1 & \textbf{72.3} & 88.2 & \textbf{44.18} (100\%) \\
SWA & ws=5120 & 59.4/29.3/31.7 & 34.6/8.1/20.1 & 33.8/9.4/22.5 & \textbf{41.9} & 17.2 & 61.9 & \textbf{88.6} & 40.68 (92.1\%) \\
\cellcolor[rgb]{0.925,0.957,1}CoMeT & \cellcolor[rgb]{0.925,0.957,1}ms=2560 & \cellcolor[rgb]{0.925,0.957,1}\textbf{63.8}/32.5/\textbf{34.3} & \cellcolor[rgb]{0.925,0.957,1}\textbf{37.3}/\textbf{9.9}/\textbf{21.6} & \cellcolor[rgb]{0.925,0.957,1}36.7/12.3/24.7 & \cellcolor[rgb]{0.925,0.957,1}37.7 & \cellcolor[rgb]{0.925,0.957,1}\textbf{24.4} & \cellcolor[rgb]{0.925,0.957,1}67.5 & \cellcolor[rgb]{0.925,0.957,1}\textbf{88.6} & \cellcolor[rgb]{0.925,0.957,1}43.14 (97.6\%) \\
\cdashline{1-10}
\specialrule{0em}{3pt}{1pt}
\multicolumn{10}{l}{\textit{Llama-3.1-8B-Instruct}} \\
\specialrule{0em}{1pt}{1pt}
Full Attn (FT) & Full Context & 61.3/\textbf{32.1}/\textbf{33.4} & 33.4/7.6/19.3 & \textbf{36.6}/\textbf{12.3}/\textbf{24.6} & \textbf{37.0} & 17.4 & 51.3 & 86.0 & 38.76 (100\%) \\
SWA & ws=5120 & 60.2/31.0/32.2 & 33.3/8.1/19.6 & 34.0/10.7/22.6 & 36.3 & 15.9 & 44.0 & 86.3 & 37.04 (95.6\%) \\
\cellcolor[rgb]{0.925,0.957,1}CoMeT & \cellcolor[rgb]{0.925,0.957,1}ms=2560 & \cellcolor[rgb]{0.925,0.957,1}\textbf{62.9}/31.5/\textbf{33.4} & \cellcolor[rgb]{0.925,0.957,1}\textbf{36.6}/\textbf{9.5}/\textbf{21.1} & \cellcolor[rgb]{0.925,0.957,1}36.5/11.3/23.9 & \cellcolor[rgb]{0.925,0.957,1}36.9 & \cellcolor[rgb]{0.925,0.957,1}\textbf{23.6} & \cellcolor[rgb]{0.925,0.957,1}\textbf{54.7} & \cellcolor[rgb]{0.925,0.957,1}\textbf{87.3} & \cellcolor[rgb]{0.925,0.957,1}\textbf{40.55} (104.6\%) \\
\bottomrule
\end{tabular}
\caption{Generalization results on \scrolls{} benchmark across model families and scales. CoMeT consistently outperforms SWA across all settings. The percentage in parentheses denotes performance relative to the fine-tuned full attention baseline. CoMeT's relative performance improves with model scale (95.0\% on Qwen3-4B vs.\ 97.6\% on Qwen3-14B), and on Llama-3.1-8B-Instruct, CoMeT surpasses the full attention baseline by 4.6\%.}
\label{tab:model_generalization}
\end{table*}

\begin{figure*}[t]
\centering
\captionsetup[sub]{font=small, labelfont=small}
 \begin{minipage}{0.25\textwidth}
        \centering
        \subcaptionbox{Prefill latency\label{fig:prefill_time}}{
         \includegraphics[width=\linewidth]{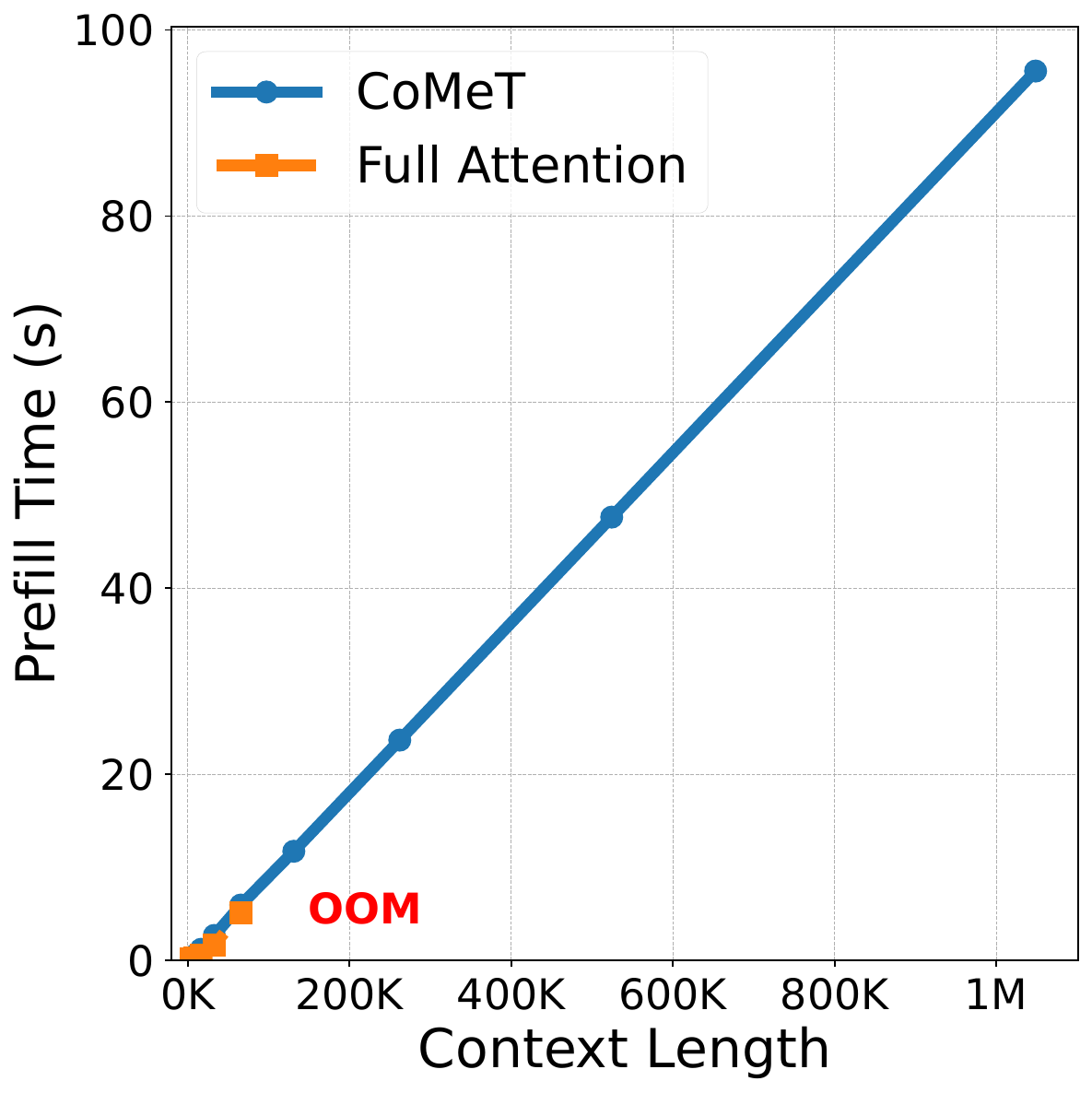}
        }
 \end{minipage}%
 \begin{minipage}{0.25\textwidth}
        \centering
        \subcaptionbox{Prefill memory usage\label{fig:prefill_memory}}{
         \includegraphics[width=\linewidth]{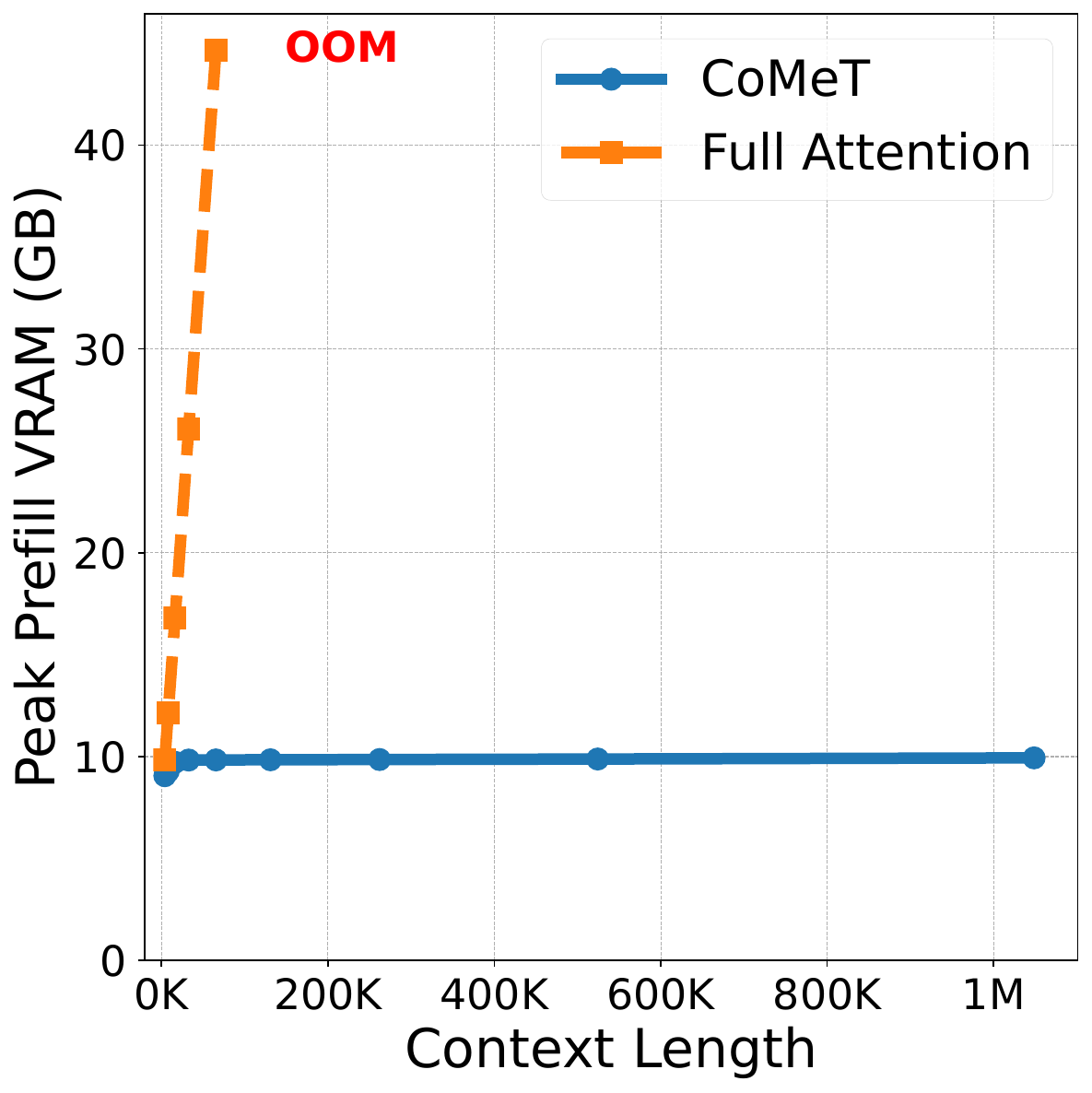}
        }
 \end{minipage}%
 \begin{minipage}{0.25\textwidth}
    \centering
    \subcaptionbox{Per-token decode latency\label{fig:decode_time}}{
    \includegraphics[width=\linewidth]{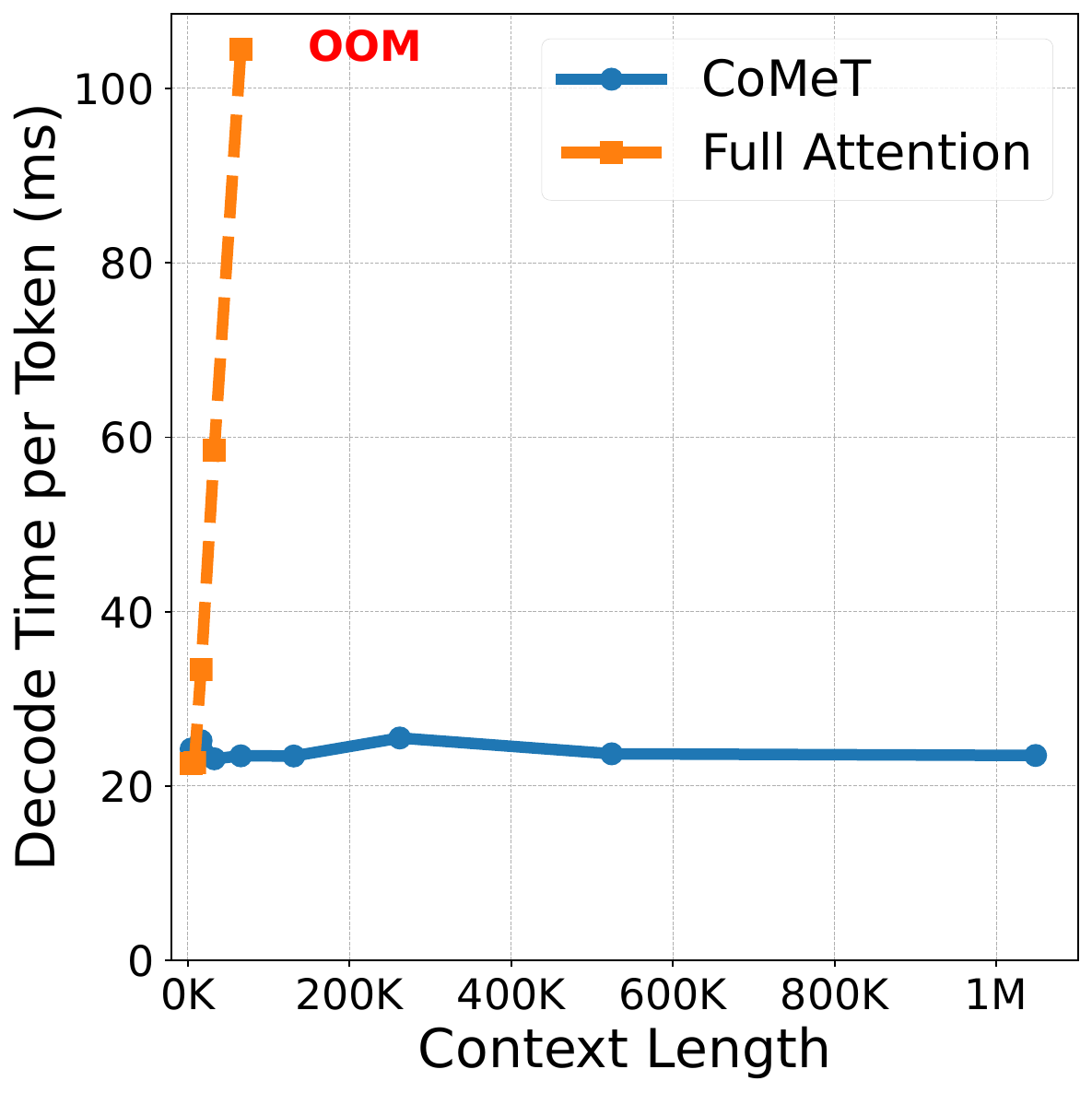}
    }
\end{minipage}%
\begin{minipage}{0.25\textwidth}
    \centering
    \subcaptionbox{Decode memory usage\label{fig:decode_memory}}{
    \includegraphics[width=\linewidth]{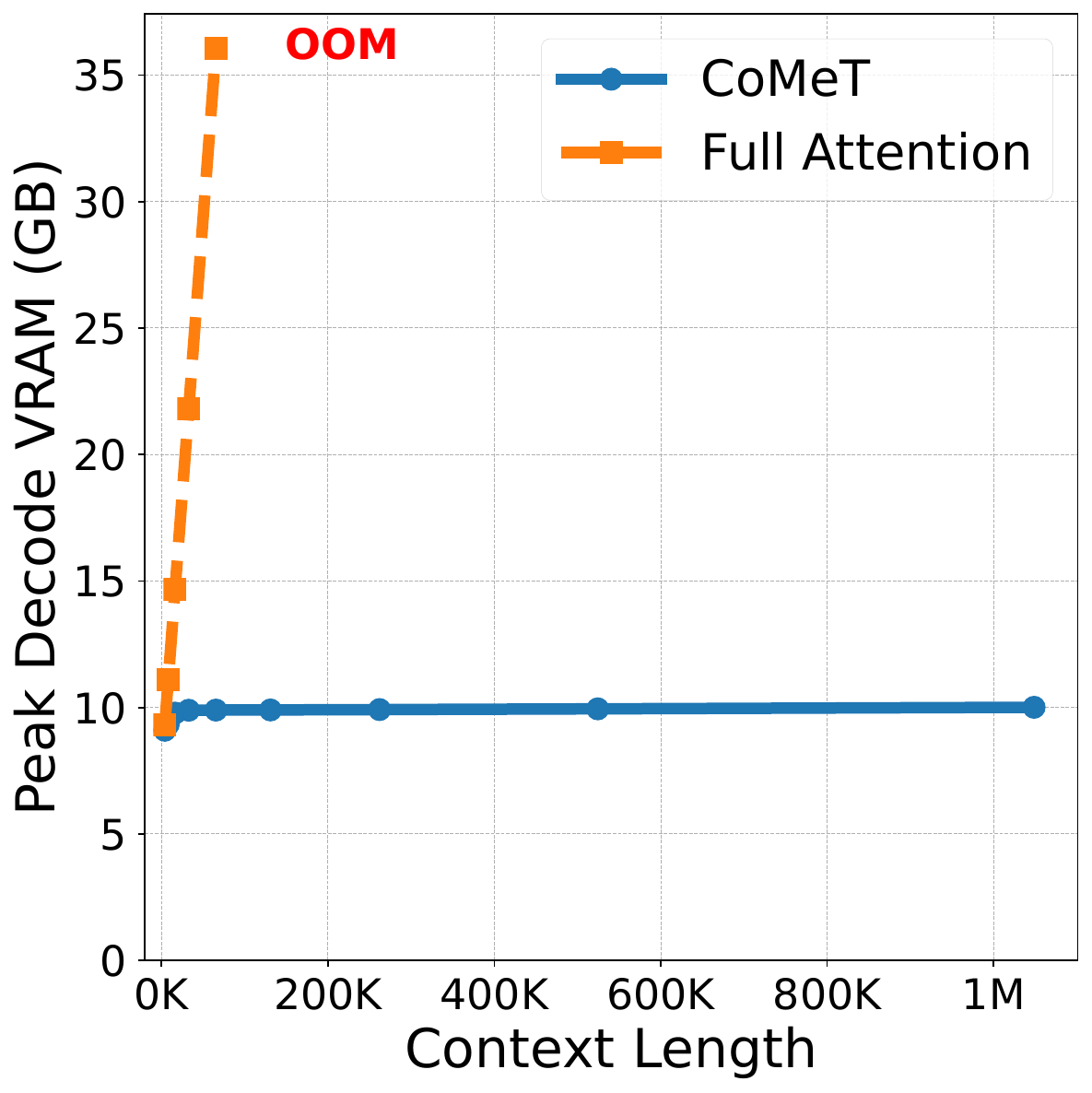}
    }
\end{minipage}
\caption{Performance comparison of prefill and decode phases. (a) and (b) show the latency and memory usage during the prefill phase, while (c) and (d) present the per-token latency and memory consumption during the decode phase. Notably, the experiment is capped for Full Attention at 128k due to an OOM error, while our method (CoMeT) demonstrates scalability up to 1M.}
\label{fig:performance_comparison}
\end{figure*}

\subsection{Generalization Across Model Families and Scales}
To validate CoMeT's plug-and-play nature across diverse architectures and scales, we evaluate it on two additional models: Qwen3-14B and Llama-3.1-8B-Instruct. Results on the \scrolls{} benchmark are presented in Table~\ref{tab:model_generalization}.

CoMeT consistently outperforms SWA across all three models, demonstrating broad generalizability. Notably, CoMeT's performance relative to full attention improves as model scale increases, retaining 95.0\% of full attention performance on Qwen3-4B but 97.6\% on Qwen3-14B. Furthermore, on the Llama-3.1-8B-Instruct model, CoMeT surpasses the full-attention baseline by 4.6\%, suggesting that its benefits may be more pronounced in the Llama model family. These results confirm that CoMeT is an effective plug-and-play module for diverse model architectures and scales.

\subsection{Roles of Global and Temporary Memory}
\label{sec:analysis_memory_roles}

To dissect the distinct roles of our dual-memory system, we conduct ablation studies on memory allocation. We find that temporary memory is crucial for performance on in-domain sequence lengths, while global memory is paramount for extrapolation to out-of-domain lengths.

\paragraph{Temporary Memory Benefits Performance on In-Domain Lengths.} On the \scrolls{} benchmark, where tasks are within our 32k training length, temporary memory proves to be critical. As shown in Figure~\ref{fig:memory_allo}, overall performance improves with temporary memory size, saturating at 2,048 tokens. This demonstrates that temporary memory is vital for preserving the recent, detailed context. In contrast, increasing global memory offers only marginal gains here, suggesting its primary role lies elsewhere.

\paragraph{Global Memory Enables Length Extrapolation.} The gated global memory is key for handling sequences longer than the training data. On the 1M-token passkey task (Figure~\ref{fig:MTP_Fig}), a global-only memory (Figure~\ref{fig:g3072_t0}) achieves perfect accuracy. In contrast, a configuration focused on temporary memory (Figure~\ref{fig:g8_t3072}) shows degraded performance, proving less effective at preserving a single fact over extreme distances. Most tellingly, removing the gating mechanism (Figure~\ref{fig:g3072_b0_without_gate}) causes a complete performance collapse. This confirms the gate is essential for protecting key information.

\begin{table*}[ht]
\setlength{\tabcolsep}{9.5pt}
\centering
\small
\begin{tabular}{c c c c c c c c c}
\toprule
\textbf{Rank} & \textbf{GovRep} & \textbf{SumScr} & \textbf{QMSum} & \textbf{Qspr} & \textbf{Nrtv} & \textbf{QALT} & \textbf{CNLI} & \textbf{Avg} \\
\cmidrule(lr){2-2} \cmidrule(lr){3-3} \cmidrule(lr){4-4} \cmidrule(lr){5-5} \cmidrule(lr){6-6} \cmidrule(lr){7-7} \cmidrule(lr){8-8}
& R-1/2/L & R-1/2/L & R-1/2/L & F1 & F1 & F1 & EM \\
\midrule
\specialrule{0em}{1pt}{1pt}
4   & 62.4/31.3/\textbf{33.5} & 33.4/7.8/19.4 & 35.2/11.1/24.0 & 33.9 & 21.7 & 54.0 & 86.1 & 39.18 \\
\cellcolor[rgb]{0.925,0.957,1}8   & \cellcolor[rgb]{0.925,0.957,1}62.5/31.1/33.4 & \cellcolor[rgb]{0.925,0.957,1}\textbf{33.4}/\textbf{8.3}/\textbf{19.8} & \cellcolor[rgb]{0.925,0.957,1}\textbf{35.6}/\textbf{12.0}/\textbf{24.6} & \cellcolor[rgb]{0.925,0.957,1}\textbf{35.5} & \cellcolor[rgb]{0.925,0.957,1}\textbf{22.6} & \cellcolor[rgb]{0.925,0.957,1}56.0 & \cellcolor[rgb]{0.925,0.957,1}86.9 & \cellcolor[rgb]{0.925,0.957,1}\textbf{40.10} \\
64  & \textbf{62.8}/\textbf{31.4}/\textbf{33.5} & 32.5/7.7/19.1 & \textbf{35.6}/11.7/24.4 & 34.2 & 21.9 & \textbf{56.3} & \textbf{87.2} & 39.79 \\
512 & 62.6/\textbf{31.4}/33.4 & 31.2/7.0/18.7 & 35.3/11.2/24.2 & 34.2 & 20.5 & 54.2 & 86.7 & 39.02 \\
\bottomrule
\end{tabular}
\caption{Ablation study on RLA rank on the \scrolls{} benchmark. CoMeT (highlighted) uses $r=8$ by default.}
\label{tab:rank_ablation}
\end{table*}

\subsection{Efficiency Analysis}

We conduct an in-depth analysis of CoMeT's time and space efficiency during inference, comparing it with the Full Attention setting. Figure~\ref{fig:performance_comparison} presents the results based on the Qwen3-4B-Instruct-2507 model.
CoMeT demonstrates superior space efficiency, maintaining constant peak memory consumption of $\sim10$GB regardless of context length (Figures~\ref{fig:prefill_memory} and~\ref{fig:decode_memory}), while Full Attention's memory usage grows linearly, reaching OOM at 128k tokens. In terms of time efficiency, CoMeT's prefill latency scales linearly with context length (Figure~\ref{fig:prefill_time}), and its per-token decoding latency remains stable at $\sim22$ms (Figure~\ref{fig:decode_time}). In contrast, Full Attention's decoding latency increases linearly, reaching 104ms at 65k tokens.
Additional experiments with a smaller model further validate the theoretical complexity differences: CoMeT maintains linear prefill latency and constant memory consumption, while Full Attention shows quadratic growth in prefill latency and linear growth in peak memory (Figures~\ref{sub2} and~\ref{sub3}).
These results demonstrate CoMeT's significant efficiency advantages in processing long contexts. For a more detailed analysis, please refer to Appendix~\ref{app:efficiency_analysis} 
and Appendix~\ref{app:rwkv_efficiency}.

\subsection{Sensitivity to Hyperparameters}
Table~\ref{tab:rank_ablation} presents an ablation study on the RLA rank. Performance is largely stable across ranks from 4 to 512, with rank 8 achieving the best average score of 40.10. Increasing the rank to 512 leads to slight performance degradation. We therefore use $r=8$ as the default setting. Other hyperparameters are determined based on our preliminary experiments, 
with detailed results provided in Appendix~\ref{app:hyperparameter_ablation}.

\subsection{Gating Value Visualization}

\begin{figure}[t]
\centering
\captionsetup[sub]{font=small, labelfont=small}
 \begin{minipage}{0.215\textwidth}
        \centering
        \subcaptionbox{Layer 33\label{layer_33}}{
         \includegraphics[width=\linewidth]{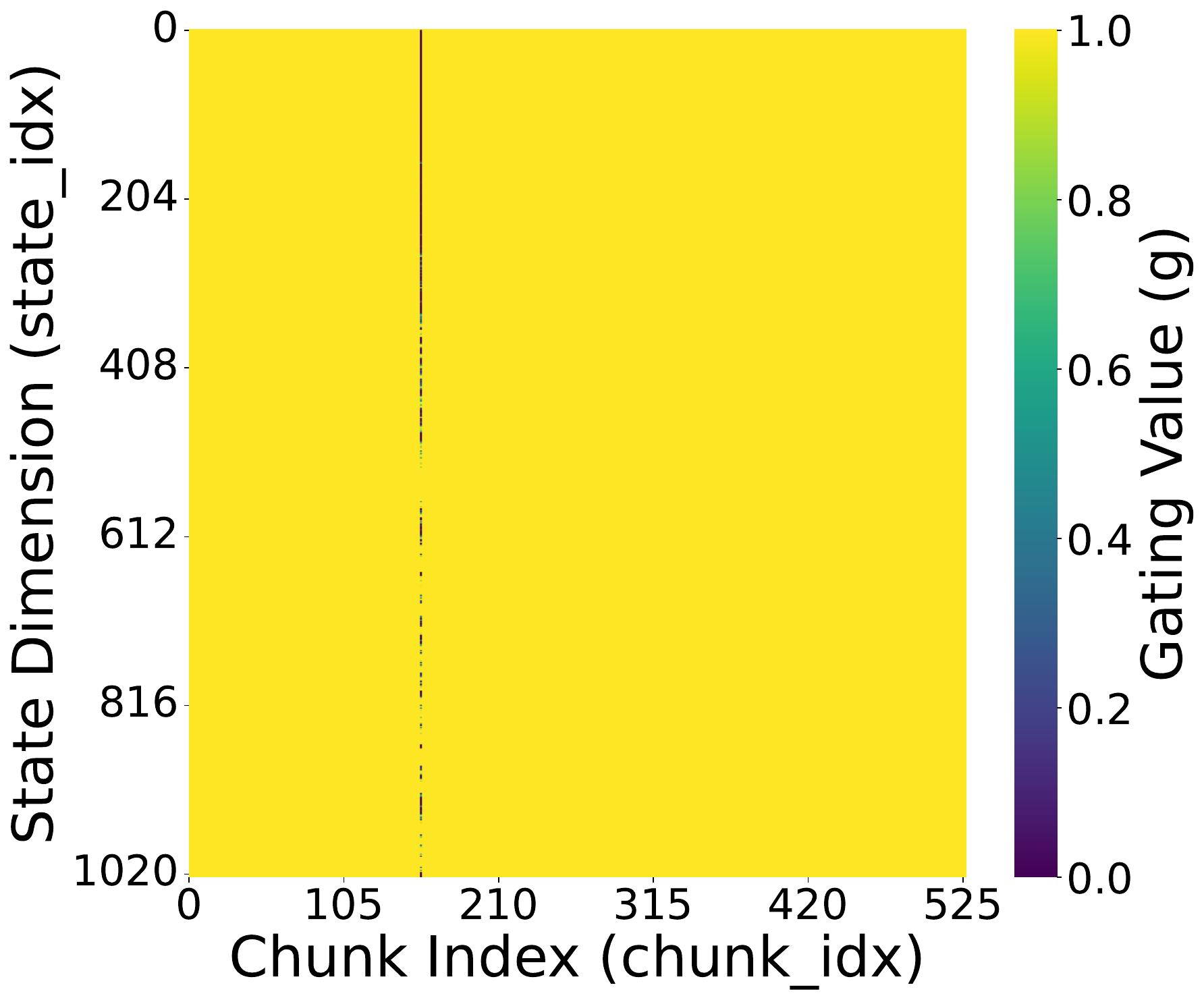} 
        }
 \end{minipage}
 \qquad
 \begin{minipage}{0.215\textwidth}
    \centering
    \subcaptionbox{Layer 7\label{layer_7}}{ 
    \includegraphics[width=\linewidth]{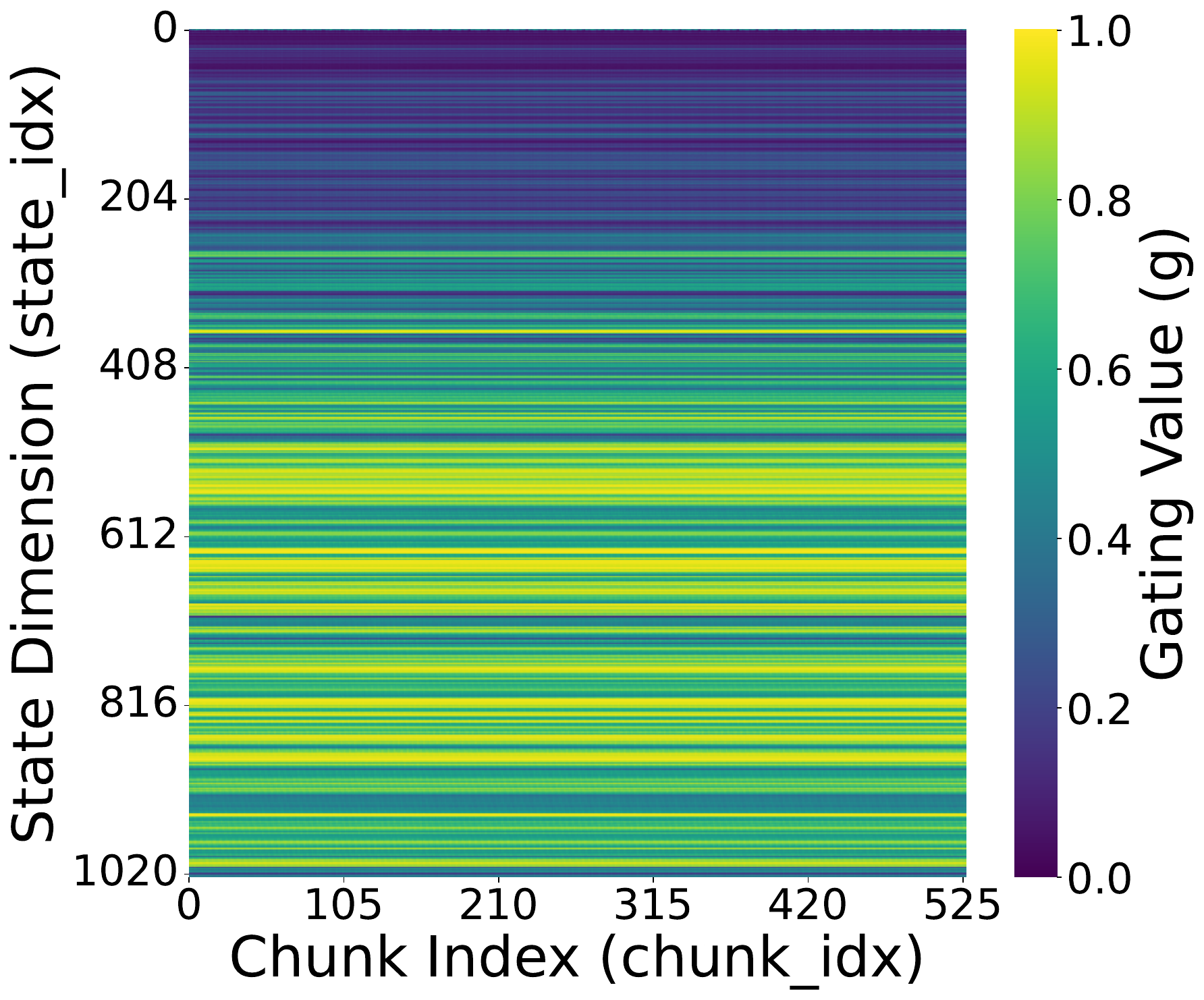}
    }
\end{minipage}
\caption{Visualization of gating values when processing a 1M-token passkey retrieval task, where the passkey appears at chunk 157 (30\% depth). The x-axis represents chunk indices and the y-axis represents the IDs of 1024 global memory states. (a) Layer 33 shows gate values dropping to 0 at chunk 157 when encountering the passkey, then consistently remaining at 1 to preserve the critical information. (b) Layer 7 exhibits differentiated gating patterns across states, indicating varied forgetting rates and multi-scale memory preservation.}
\label{fig:gating_vis}
\end{figure}

To gain deeper insights into the role of the gating mechanism, we conduct a visualization analysis of CoMeT's behavior when processing extremely long texts. We select a 1M-token passkey retrieval task where the key is inserted at 30\% depth. The analysis reveals that the gate is crucial for long-term retention, particularly in the model's deeper layers (e.g., 24, 28, 29, and 33). As illustrated in Figure~\ref{layer_33}, upon encountering the passkey, the gate values in layer 33 drop to 0, allowing the critical information to be written into the global state. Subsequently, the gates close (values remain at 1), effectively shielding this information from being overwritten by later chunks. In contrast, other layers exhibit more nuanced behavior. Figure~\ref{layer_7} shows that different states within the same layer have varied gating patterns, suggesting they possess differentiated forgetting rates. This allows the model to preserve information across multiple time scales. Complete visualization results for all layers are provided in Appendix~\ref{passkey_vis}.

\section{Conclusion}

In this work, we introduce CoMeT, a novel plug-in module that overcomes the scaling limitations of standard Transformers. By combining gated global memory for long-term dependencies with temporary FIFO memory for recent details, CoMeT achieves constant memory usage and linear time complexity. Remarkably, CoMeT trained on 32k contexts accurately retrieves information from 1M token sequences with $21\times$ speedup over full attention. Combined with strong performance on the \scrolls{} benchmark and proven real-world utility, CoMeT makes arbitrarily long-context processing practical for LLMs.

\section*{Limitations}

While CoMeT effectively coordinates global and temporary memory, our current framework has not yet explored integration with episodic memory (test-time training) and external memory (such as notebooks and RAG-based knowledge bases). These components play crucial roles in human cognition for complex tasks. We view these not as fundamental flaws but as exciting avenues for future research. CoMeT's modular architecture provides a natural foundation for incorporating these additional memory types, and we hope our work will inspire further exploration in this direction.

\section*{Ethical Considerations and Societal Impact}
CoMeT advances efficient long-context processing by significantly reducing memory and computational requirements, which can democratize access to large language models for researchers and practitioners with limited hardware resources. By enabling effective processing of very long sequences on modest infrastructure, CoMeT has the potential to facilitate personalized applications---such as user behavior understanding and intelligent assistants---that were previously accessible only to large organizations with substantial compute budgets.

However, the same capabilities that make CoMeT beneficial also warrant careful consideration of potential risks. The ability to efficiently process very long sequences of user interaction logs, as demonstrated in our UQA experiments, could be misused for large-scale surveillance or unauthorized analysis of personal behavior data. We encourage practitioners deploying CoMeT in such contexts to adhere to applicable data privacy regulations and to implement appropriate safeguards to protect user privacy.

\section*{Acknowledgments}
This work was supported by Alibaba Group through Alibaba Research Intern Program, the National Natural Science Foundation of China (Nos. U24A20334 and 62276056), the Yunnan Fundamental Research Projects (No.202401BC070021), the Yunnan Science and Technology Major Project (No. 202502AD080014), the Fundamental Research Funds for the Central Universities (Nos. N25BSS054 and N25BSS094), and the Program of Introducing Talents of Discipline to Universities, Plan 111 (No.B16009).




\bibliography{main}

\clearpage
\appendix

\section{Pipeline Bubble Visualization}
\label{appendix:bubble}

\begin{figure*}[t]
    \centering
    \includegraphics[width=1\linewidth]{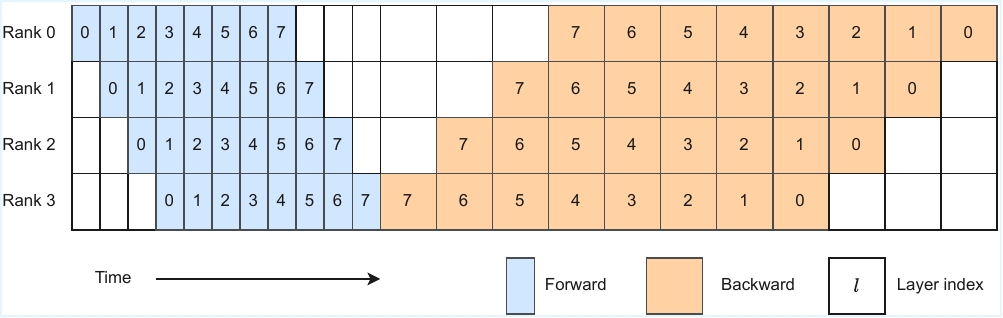}
    \caption{The visualization of pipeline bubbles for Layer-Level Pipeline Parallelism strategies. A sequence is divided into 4 segments (one per GPU) and each GPU has an 8‑layer model. Blue cells indicate forward computation and orange cells indicate backward computation; white cells represent idle (bubble) time. The layer index $l$ is shown within each cell. Compared to the naive strategy, our layer-level pipeline significantly reduces the bubble ratio.}
    \label{fig:bubbles}
\end{figure*}

To complement the discussion in Section~\ref{sec:pp}, we provide a detailed visualization of pipeline bubbles for our proposed Layer-Level Pipeline Parallelism strategies. Figure~\ref{fig:bubbles} illustrates the execution timeline across four GPU ranks, where a sequence is divided into 4 segments\footnote{If a segment is too long, it is further split into multiple chunks with chunk size within a GPU.} (one per GPU) and each GPU has an 8‑layer model.

\section{Experimental Setup and Baseline Configurations}

\label{sec:setup}

In our experimental setup, we employ Qwen3-4B-Instruct-2507~\cite{qwen3technicalreport} as the backbone model by default. We uniformly set the memory budget to $\sim3072$ tokens for all baseline methods, except for the full attention which serves as the performance upper bound. Specifically, for text-level compression methods, we compress texts to 3,072 tokens, while texts shorter than 3,072 tokens remain uncompressed. For activation-level compression methods, given that the model is trained on sequences of 32k length, we set the chunk size to 2048 and employ 192 special tokens for compression per chunk. The configurations for other baseline methods are as follows: SWA adopts a window size of 5,120, Transformer-XL retains the most recent 3072 hidden states with a chunk size of 2,048, and HMT uses 32 sensory memory slots with a long-term memory budget of 3040. We configure CoMeT with 512 global memory, 2,048 temporary memory, a chunk size of 2,048, and insert one compression token every 8 context tokens, as this configuration achieves excellent performance without requiring the larger budget used by other baseline methods, as demonstrated in Table~\ref{fig:memory_allo}.

Unless otherwise noted, we adopt a unified training configuration for all methods requiring fine-tuning: batch size of 64 with sequences of varying lengths packed to 32k for training; learning rate of $5e-5$ with 10 warmup steps followed by cosine decay to 0; Adam optimizer with hyperparameters $\beta_1=0.9$ and $\beta_2=0.999$. For training-free methods, we directly evaluate their performance on the fine-tuned full attention model to assess the effectiveness of pure compression strategies.

\section{Dataset Construction Details}
\label{app:dataset_details}

This appendix provides additional information on the construction of the mixed datasets used for fine-tuning our models, as mentioned in the main experiments section.

\paragraph{\scrolls{} Mixed Dataset.} 
To comprehensively evaluate the long-context processing capabilities of our models, we create a unified training and validation dataset derived from the \scrolls{} benchmark~\cite{shaham-etal-2022-scrolls}. This dataset amalgamates samples from all seven constituent tasks of \scrolls{}: \textit{GovReport}, \textit{SummScreenFD}, \textit{QMSum}, \textit{Qasper}, \textit{NarrativeQA}, \textit{QuALITY}, and \textit{ContractNLI}. 

To manage training constraints and focus on a long-but-tractable context window, we filter the combined dataset to include only those examples where the total input sequence length does not exceed 32,768 tokens. The final dataset comprises 41,496 training samples and 7,455 validation samples. This process ensures that our training data is diverse, covering a wide range of tasks (summarization, question answering, natural language inference) and domains, while remaining within the specified maximum length for our fine-tuning process. During training, these variable-length sequences are packed into batches with a fixed total length of 32k tokens to maximize computational efficiency. 

\paragraph{Shorter-Context QA Mixed Dataset.}
To ensure that our model's long-context adaptations do not degrade its performance on shorter sequences, we also construct a separate training set from established multi-hop Question Answering (QA) benchmarks. This dataset is created by sampling 20,000 examples from the \textit{2WikiMQA} dataset and another 20,000 examples from the \textit{HotpotQA} dataset. These 40,000 samples are then mixed to form a unified training set. Training on this mixed dataset allows the model to maintain its proficiency on tasks that require reasoning over shorter, more concise contexts, demonstrating that the CoMeT architecture does not compromise performance on standard-length inputs.

\paragraph{UQA Dataset.}
This dataset originates from a proprietary collection of user interaction logs from a major online e-commerce entity, which remains anonymous for confidentiality purposes. The core objective is to assess a model's ability to comprehend and reason over extended user activity sequences. The tasks designed for this dataset are diverse and include: (1) forecasting user interest in new product categories from their clickstream data; (2) providing tailored product recommendations informed by both click history and explicit search queries; and (3) synthesizing user behavior into a coherent summary. We employ an LLM-as-judge evaluation methodology, utilizing Qwen2.5-72B-Instruct as the judge model to assess the quality of model responses.

\paragraph{Long-Horizon Agent Dataset.}
This dataset is curated through a multi-stage, expert-driven methodology. Initially, we conduct a systematic analysis of GitHub issue forums to identify recurring and complex real-world software engineering problems. Subsequently, domain experts formulate a suite of tasks designed to emulate these challenges. Training trajectories are then generated by capturing the step-by-step interactions of an expert-operated, state-of-the-art agent model as it works to resolve these tasks.



\section{Extrapolation on Long-Context \scrolls{} Subset}
\label{app:extrapolation}

To evaluate CoMeT's extrapolation ability on realistic tasks, we tested our model 
(fine-tuned on 32k contexts) on a subset of the SCROLLS validation set containing 
only sequences \textbf{longer than 32k tokens}. This subset comprises 4,854 samples 
with an average length of 91k and a maximum length of 484k.

As shown in Table~\ref{tab:extrapolation}, CoMeT substantially outperforms Sliding 
Window Attention (SWA) on both tasks, with a particularly large gain on NrtvQA 
(13.2 vs.\ 5.8, a 128\% improvement). This confirms that CoMeT's memory module 
continues to function effectively on sequences well beyond its training length. 
While CoMeT does not match Full Attention (FA) on this subset, it is worth noting 
that the base model (Qwen-3.4B-Instruct) natively supports a 256k context length, 
giving FA a significant advantage. Training CoMeT on longer contexts (e.g., 128k) 
via its layer-level pipeline parallelism may further close this gap.

\begin{table}[h]
\centering
\footnotesize
\begin{tabular}{lcc}
\toprule
\textbf{Method} & \textbf{GovRep} & \textbf{NrtvQA} \\
\cmidrule(lr){2-2} \cmidrule(lr){3-3}
& R-1/2/L & F1  \\
\midrule
FA (Full Context) & \textbf{48.8} / \textbf{23.7} / \textbf{26.3} & \textbf{20.4} \\
SWA (ws=5120)      & 31.5 / 10.5 / 18.4 & 5.8  \\
\cellcolor[rgb]{0.925,0.957,1}\textbf{CoMeT} (ms=2560) & \cellcolor[rgb]{0.925,0.957,1}38.6 / 14.1 / 19.9 & \cellcolor[rgb]{0.925,0.957,1}13.2 \\
\bottomrule
\end{tabular}
\caption{Performance on the long-context subset of \scrolls{} (input $>$ 32k tokens; 
4,854 samples, avg.\ 91k, max 484k). CoMeT is fine-tuned only on 32k contexts.\label{tab:extrapolation}}
\end{table}

\section{Passkey Retrieval Task}
\label{sec:passkey}
Below we showed the chatML format of the passkey task.

\begin{quote}
[{"role": "system", "content": "You are an expert at finding a specific 'pass key' inside a long document. When asked, you will reply with only the pass key and nothing else."}, {"role": "user", "content": " There is an important info hidden inside a lot of irrelevant text. Find it and memorize them. I will quiz you about the important information there. To bake a cake, you need flour, sugar, and eggs. Mix them well. Bake at 350 degrees. \underline{(repeat x times)} The pass key is \textbf{1392093}. Remember it. \textbf{1392093} is the pass key. To bake a cake, you need flour, sugar, and eggs. Mix them well. Bake at 350 degrees. \underline{(repeat y times)} What is the pass key? The pass key is"}]
\end{quote}

\section{Detailed Efficiency Analysis}
\label{app:efficiency_analysis}

This section provides an in-depth analysis of the time and space efficiency of CoMeT during inference. We conduct experimental comparisons between the standard Full Attention architecture and the modified CoMeT architecture, focusing on system overhead during both the prefill and decode phases under varying context lengths. Figure~\ref{fig:performance_comparison} presents detailed comparative results based on the Qwen3-4B-Instruct-2507 model.

\paragraph{Space Efficiency.} As illustrated in Figures~\ref{fig:prefill_memory} and~\ref{fig:decode_memory}, CoMeT demonstrates remarkably superior efficiency. In both the prefill and decode phases, CoMeT maintains a constant peak memory consumption of approximately 10GB, remaining completely unaffected by context length growth. In contrast, Full Attention exhibits linear memory growth with increasing sequence length, encountering out-of-memory errors when processing 128k context tokens. This demonstrates that CoMeT's constant space complexity enables it to handle sequences of arbitrary length.

\paragraph{Time Efficiency.} During the prefill phase (Figure~\ref{fig:prefill_time}), CoMeT's latency scales linearly with context length, which aligns with its chunk-by-chunk processing mechanism. The advantage of CoMeT becomes even more pronounced in the decode phase. As shown in Figure~\ref{fig:decode_time}, the per-token decoding latency remains consistently stable at around 22ms regardless of context length. Conversely, Full Attention's decoding latency increases linearly with growing context, reaching 104ms at 65k tokens, nearly 5 times that of CoMeT, with this gap continuing to widen as sequence length increases.

To more clearly demonstrate the asymptotic complexity differences between the two architectures for longer sequences, we conduct supplementary experiments using a smaller model ($d_{model}=768$, 12 layers). As shown in Figures~\ref{sub2} and~\ref{sub3}, the experimental results clearly validate our theoretical analysis: Full Attention exhibits quadratic growth in prefill latency and linear growth in peak memory, while CoMeT maintains linear prefill latency and constant memory consumption. Taken together, these experimental results convincingly demonstrate that CoMeT possesses overwhelming efficiency advantages in both time and space when processing long contexts, making it a robust solution for efficient long-sequence processing.

\section{Efficiency Comparison with RWKV}
\label{app:rwkv_efficiency}

We benchmark CoMeT's latency against RWKV-7-Goose-World-2.9B-HF using the FLA 
library and the official Hugging Face implementation. Note that RWKV7 could not 
be tested beyond a context length of 32,768 due to a Triton/CUDA kernel error.
Table~\ref{tab:rwkv_latency} presents the results.

\begin{table}[htbp]
\setlength{\tabcolsep}{7.5pt}
\centering
\small
\begin{tabular}{l c c c c}
\toprule
\textbf{Model} & \makecell[l]{\textbf{Input}\\\textbf{Length}} & \makecell[l]{\textbf{Prefill}\\\textbf{(ms)}} & \makecell[l]{\textbf{Mem }\\\textbf{(MB)}} & \makecell[l]{\textbf{decode}\\\textbf{(ms)}} \\
\midrule
RWKV7  & 4096    & 136.8   & 7614  & 35.42 \\
CoMeT  & 4096    & 235.3   & 9280  & 23.25 \\
\midrule
RWKV7  & 8192    & 282.6   & 9555  & 34.96 \\
CoMeT  & 8192    & 519.3   & 9504  & 22.10 \\
\midrule
RWKV7  & 16384   & 375.2   & 13435 & 34.85 \\
CoMeT  & 16384   & 1184.0  & 9954  & 24.20 \\
\midrule
RWKV7  & 32768   & 874.8   & 21195 & 34.58 \\
CoMeT  & 32768   & 2678.9  & 10068 & 22.10 \\
\midrule
CoMeT  & 65536   & 5943.0  & 10074 & 22.50 \\
\midrule
$\cdots$ & $\cdots$ & $\cdots$ & $\cdots$ & $\cdots$ \\
\midrule
CoMeT  & 1048576 & 95574.7 & 10176 & 22.50 \\
\bottomrule
\end{tabular}
\caption{Latency comparison between CoMeT and RWKV7 at varying input lengths.}
\label{tab:rwkv_latency}
\end{table}

Two observations emerge from these results. First, both RWKV7 and CoMeT achieve 
\textbf{constant per-token decoding time} regardless of input length: CoMeT 
stabilizes at $\sim$22ms while RWKV7 stabilizes at $\sim$34ms. Second, CoMeT's 
\textbf{peak memory usage remains essentially constant} as context grows, whereas 
RWKV7's memory footprint increases with input length. We attribute RWKV7's faster 
prefill speed and its growing memory consumption to the parallel algorithm it 
employs during the prefill stage.

\section{Sensitivity to Hyperparameters}
\label{app:hyperparameter_ablation}

We conduct initial ablation studies on a different set of QA tasks to select 
our main hyperparameters before the \scrolls{} experiments.

\paragraph{Chunk Size and Compression Token Interval.}
Table~\ref{tab:hyperparam_ablation} reports results across varying chunk sizes, 
temporal strides, and temporal budgets. The key findings are:
\begin{enumerate}
    \item A larger chunk size (\emph{e.g.}, 2048) generally yields better 
    performance, but with diminishing returns and reduced efficiency. 
    We chose 2048 as a balance.
    \item A compression token interval of 8 (covering the last $\sim$16k tokens 
    in temporary memory) performed better than an interval of 16 (covering $\sim$32k).
\end{enumerate}

\begin{table*}[htbp]
\setlength{\tabcolsep}{4pt}
\centering
\small
\begin{tabular}{c c c c c c c c c c}
\toprule
\textbf{Chunk} & \textbf{Global} & \textbf{Temp} & \textbf{Temp} 
& \textbf{2WikiMQA} & \textbf{HotpotQA} & \textbf{NarrativeQA 32K} & \textbf{NarrativeQA 64K} & \textbf{NQ} \\
\textbf{Size}& \textbf{Memory}& \textbf{Stride}& \textbf{Budget}  & EM/F1 & EM/F1 & EM/F1 & EM/F1 & EM/F1 \\
\midrule
\specialrule{0em}{1pt}{1pt}
512  & 512 & 16 & 2048 & 64.4/71.1 & 55.9/71.3 & 12.8/32.3 & 12.4/27.8 & 68.3/78.0 \\
512  & 512 & 8  & 2048 & 67.7/74.3 & 58.5/73.6 & 13.7/32.0 & 12.1/27.1 & 67.7/77.8 \\
1024 & 512 & 16 & 2048 & 69.0/75.5 & 58.8/74.2 & 13.9/33.2 & 12.7/28.1 & 70.7/79.8 \\
1024 & 512 & 8  & 2048 & 73.7/79.7 & 61.0/76.4 & \textbf{15.5}/34.3 & 12.2/27.7 & 70.4/79.3 \\
1024 & 512 & 8  & 4096 & 73.1/79.2 & 59.9/75.6 & 15.4/34.0 & 13.9/29.7 & 69.1/78.4 \\
2048 & 512 & 8  & 2048 & 75.5/81.0 & \textbf{65.9}/80.0 & \textbf{15.5}/33.8 & 12.1/27.6 & 73.9/82.4 \\
2048 & 512 & 8  & 4096 & \textbf{76.6}/\textbf{82.1} & 65.8/\textbf{80.1} & 15.3/34.5 & 12.7/28.7 & 73.2/82.2 \\
2048 & 512 & 16 & 4096 & 73.3/79.1 & 65.4/79.8 & 13.6/32.2 & 12.2/27.0 & 71.6/80.2 \\
\midrule
Full Attn & -- & -- & -- & 75.4/80.8 & 65.0/78.9 & 15.4/\textbf{35.5} & \textbf{15.9}/\textbf{32.0} & \textbf{77.3}/\textbf{84.2} \\
\bottomrule
\end{tabular}
\caption{Ablation on chunk size and compression token interval across QA tasks. 
Bold denotes the best result. This NarrativeQA is not from the 
\scrolls{} dataset, hence the different results.}
\label{tab:hyperparam_ablation}
\end{table*}

\section{Gating Value Visualization for All Layers}
\label{passkey_vis}

\begin{figure*}
    \centering
    \includegraphics[width=1\linewidth]{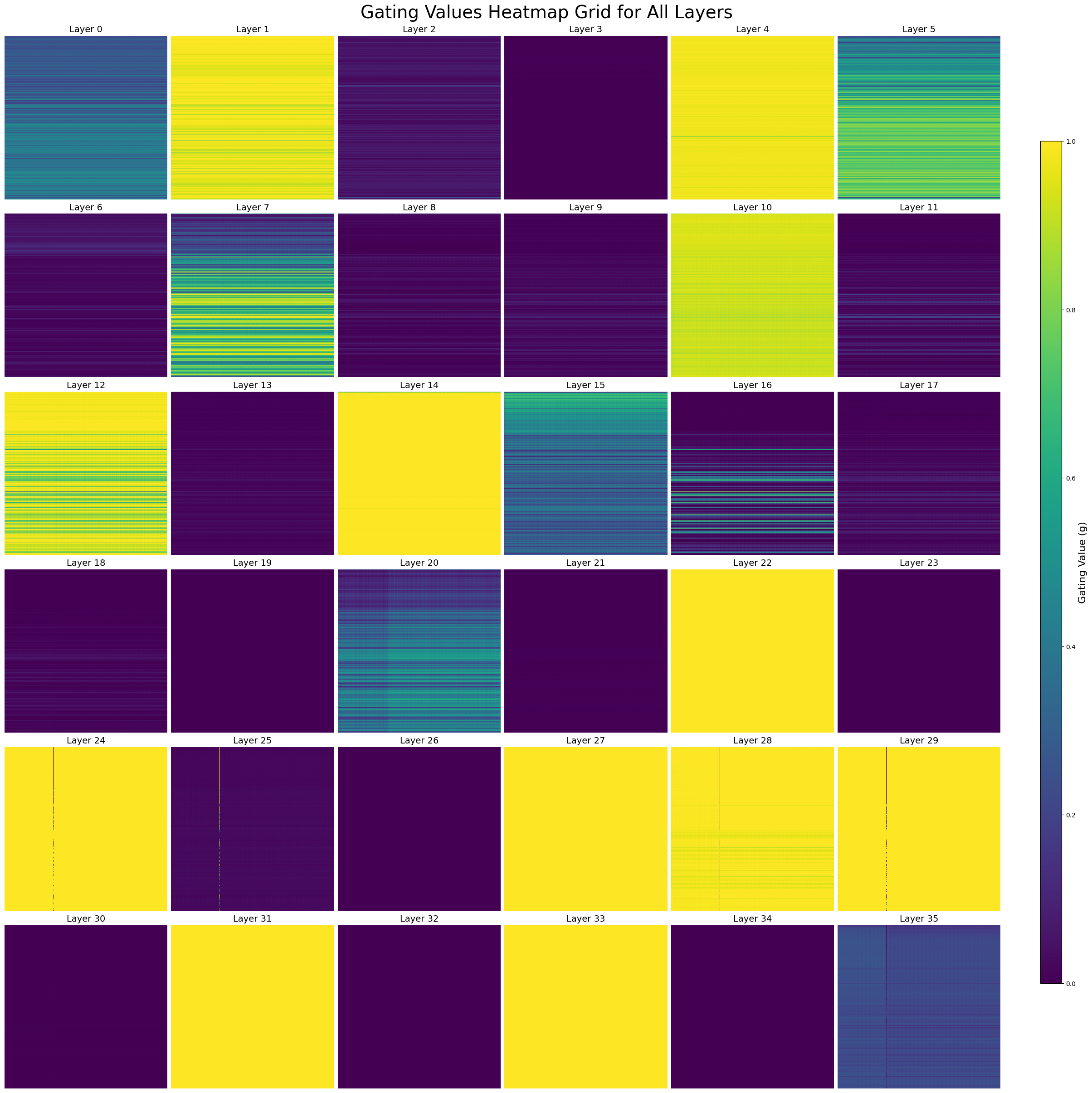}
    \caption{Complete visualization of gating values across all 36 layers when processing the 1M-token passkey retrieval task. Each subplot shows the gating heatmap for a specific layer, with the x-axis representing chunk indices and the y-axis representing the IDs of 1024 global memory states. The passkey appears at chunk 157 (30\% depth).}
    \label{fig:all_layers_gating}
\end{figure*}

For completeness, we provide a comprehensive visualization of the gating values across all layers of CoMeT when processing the 1M-token passkey retrieval task (with the passkey inserted at 30\% depth). Figure~\ref{fig:all_layers_gating} presents the gating heatmaps for all 36 layers of the Qwen3-4B model. 

\section{Analysis of Failure Cases and Information Loss}
\label{app:failure_analysis}

Following the qualitative analysis of failure cases on the \textbf{Qasper} dataset, 
where CoMeT showed a noticeable performance gap compared to the full attention 
baseline with the Qwen3-4B-Instruct-2507 model, we present representative failure 
patterns in Table~\ref{tab:failure_cases}.

\begin{table*}[htbp]
\setlength{\tabcolsep}{5pt}
\centering
\small
\begin{tabular}{p{1.0cm} p{2.0cm} p{2.0cm} p{2.0cm} p{2.0cm} p{4.5cm}}
\toprule
\makecell[l]{\textbf{Case}\\\textbf{ID}} & \textbf{Question} & \makecell[l]{\textbf{CoMeT's}\\\textbf{Output}} & \makecell[l]{\textbf{Full Attn's}\\\textbf{Output}} & \makecell[l]{\textbf{Correct}\\\textbf{Answer}} & \textbf{Failure Analysis} \\
\midrule
\specialrule{0em}{1pt}{1pt}
0416... & Is this hashtag prediction task an established task, or something new? & ``Yes'' & ``Unanswerable'' & ``established task'' & \textbf{Detail Loss \& Oversimplification:} CoMeT correctly identified the affirmative nature of the answer but failed to extract the more informative core phrase, ``established task.'' It provided the simplest possible correct response (``Yes''). This suggests its memory compression prioritized the binary judgment (yes/no) while discarding the specific descriptive details. \\
\midrule
\specialrule{0em}{3pt}{1pt}
0407... & Which translation systems do they compare against? & ``a state-of-the-art phrase-based MT system'' & ``hierarchical phrase-based system BIBREF29'' & ``English-Spanish MT system'' & \textbf{Loss of Granularity \& Generalization:} The original text uses the specific term ``hierarchical phrase-based system.'' While Full Attention extracted this, CoMeT produced a generalized version. It retained the core concept (``phrase-based'') but lost the more precise modifier (``hierarchical''). This indicates CoMeT's memory mechanism may be less sensitive to specific, descriptive terms, opting for more generic concepts and resulting in a loss of information granularity. \\
\midrule
\specialrule{0em}{3pt}{1pt}
02f1a... & Why are current ELS's not sufficiently effective? & ``due to low precision rates BIBREF11 and design challenges in training datasets BIBREF12'' & ``the extracted linked entities may be too ambiguous and coarse to be considered relevant to the summary'' & ``Linked entities may be ambiguous or too common'' & \textbf{Causal Reasoning Failure:} The text establishes a causal chain: the problem is ``ambiguity and coarseness,'' which is caused by ``low precision rates...'' The question asks for the problem itself. Full Attention correctly identified the problem. CoMeT, however, skipped the problem and extracted its cause. This suggests CoMeT's compressed context flattened the causal relationship, making it unable to distinguish cause from effect when answering a ``Why'' question. \\
\midrule
\specialrule{0em}{3pt}{1pt}
00da... & What dataset is used? & ``WMT 2014 shared translation task'' & ``Europarl corpus'' & ``Europarl corpus, WMT newstest 2014, News-Commentary-11, Wikipedia from WMT 2014, Multi-UN, EU-Bookshop, Rapid, Common-Crawl (WMT 2017)'' & \textbf{Information Aggregation Failure:} The correct answer required aggregating a list of datasets from multiple, non-contiguous parts of the document. Both models failed to retrieve the complete list. This failure mode is particularly challenging for an efficient model like CoMeT. Its limited or compressed view of the global context makes it more prone to finding only the first or most locally salient answer piece, thus failing to perform the necessary long-range search and aggregation required to construct a comprehensive answer. \\
\specialrule{0em}{1pt}{1pt}
\bottomrule
\end{tabular}
\caption{Representative failure cases of CoMeT on the Qasper dataset.}
\label{tab:failure_cases}
\end{table*}

Our key finding is that while CoMeT is susceptible to these types of information 
loss, \textbf{these failure modes are not unique to our method}. The Full Attention 
baseline exhibits similar errors, especially in long-range aggregation (as in case 
\texttt{00da...}). Furthermore, the performance gap on Qasper is model-dependent: 
when using Llama-3.1-8B-Instruct, the gap between CoMeT and Full Attention on 
Qasper narrows to a negligible \textbf{0.1 F1 points} (36.9 vs.\ 37.0).

\end{document}